\documentclass[twocolumn]{svjour3}

\usepackage{multirow}
\usepackage{natbib}
\usepackage{amsmath}
\usepackage{amssymb}
\usepackage{graphicx}
\usepackage{subcaption}
\usepackage{color}
\usepackage[colorlinks]{hyperref}
\usepackage{makecell}

\title{Monocular Object Orientation Estimation using Riemannian Regression and Classification Networks}

\author{Siddharth Mahendran \and Ming Yang Lu \and Haider Ali \and Ren\'e Vidal}
\institute{S. Mahendran \and M. Lu \and R. Vidal \at Center for Imaging Science, Mathematical Institute for Data Science, Johns Hopkins University, Baltimore, MD, USA
\and 
H. Ali \at Department of Computer Science, Johns Hopkins University, Baltimore, MD, USA
\and
\email\{siddharthm, mlu21, hali, rvidal\}@jhu.edu}
\titlerunning{3D Pose Estimation using CNNs}
\authorrunning{Mahendran et al.}

\newcommand{\myparagraph}[1]{\smallskip\noindent\textbf{#1}.}
\newcommand{\expm}{\operatorname{expm}}
\newcommand{\R}{\mathcal{R}}
\newcommand{\M}{\mathcal{M}}
\newcommand{\ie}{i.e.\ }

\begin{document}
\maketitle

\begin{abstract}
We consider the task of estimating the 3D orientation of an object of known category given an image of the object and a bounding box around it. Recently, CNN-based regression and classification methods have shown significant performance improvements for this task. This paper proposes a new CNN-based approach to monocular orientation estimation that advances the state of the art in four different directions. 
First, we take into account the Riemannian structure of the orientation space when designing regression losses and nonlinear activation functions. 
Second, we propose a mixed Riemannian regression and classification framework that better handles the challenging case of nearly symmetric objects.
Third, we propose a data augmentation strategy that is specifically designed to capture changes in 3D orientation.
Fourth, our approach leads to state-of-the-art results on the PASCAL3D+ dataset.
\end{abstract}

\section{Introduction}
\label{sec:introduction}

A long-standing goal of computer vision is to teach a machine to ``see'' and understand the 3D world captured by a 2D image. One way to do this is to describe the image in terms of the objects present in it and recover the underlying geometry of the scene. This involves predicting the rigid transformations between the camera and objects in the image. This problem, known as 3D pose estimation, is an integral part of many problems in computer vision, e.g., scene understanding and reconstruction, and has recently seen renewed interest due to its applications in autonomous driving, robot manipulation and augmented reality, where the ability to reason and plan in 3D is of vital importance. 

\begin{figure*}
	\centering
	\includegraphics[width=\linewidth]{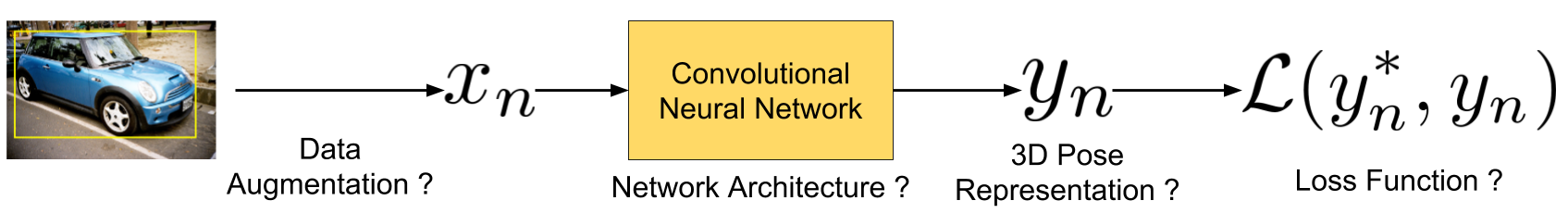}
	\caption{Overview of the problem statement: Given a 2D image $x_n$ of an object and a bounding box containing the object, design a CNN that predicts the object's 3D pose $y_n$. As part of the problem, we try to answer the following questions with geometrically meaningful components for the CNN pipeline: (i) What is an appropriate representation of 3D pose? (ii) What is a good loss function for this task? (iii) What kind of network architectures are useful for this task?? (iv) What data augmentation strategies are useful for this task?}
	\label{fig:problem_formulation}
\end{figure*}

The ``3D pose'' of an object consists of its 3D location (described by an extrinsic translation $T$) and its 3D orientation (described by an extrinsic rotation $R$). In this work, we are interested only in estimating the 3D orientation of the object. Specifically, given an image and a bounding box around an object in the image with known object category label, we consider the problem of estimating the object's orientation (see Fig.~\ref{fig:problem_formulation}). This simplification is motivated by the remarkable success of current object detection algorithms at predicting the object's scale/depth and its 2D location. Therefore, we will assume that the bounding box and category label are either given by an oracle (ground-truth) or are the output of a detection system. Throughout this paper we use the words 3D pose and orientation interchangeably.

\subsection{Prior Work}
As we will discuss in more detail in \S\ref{sec:related_work}, state-of-the-art methods for monocular object pose estimation are all based on convolutional neural networks (CNNs). One family of methods uses a CNN to extract keypoints related to the object in the image and then uses a PnP algorithm to predict the object's pose from these keypoints. Another family of methods uses a CNN to predict the pose directly from the image. Such methods follow either a regression framework, where orientation is predicted directly, or a classification framework, where the orientation space is first discretized. Our work falls in the second family of methods as we are also interested in predicting the pose directly from the input image.

In our view, a major disadvantage of existing methods is that they use representations of orientation and loss functions that do not properly take into account the geometry of the space of rotations. For example, many orientation estimation methods represent orientation in terms of Euler angles, discretize the angles into bins and solve a pose classification task while training the network with a cross-entropy loss. 
While some methods try to incorporate the cyclical symmetry of angles via a weighted cross-entropy loss, this is very different from using the Riemannian geometry of the orientation space. Other methods use better representations, but they still discretize the orientation space to eventually solve a classification problem. The drawback of classification-based approaches is that they give non-zero estimation error even with perfect pose classification accuracy due to the discretization process. Moreover, this error might be large if the binning is coarse. 

Regression-based approaches overcome errors due to binning by predicting a continuous pose. However, current methods either regress a $(\cos, \sin)$ representation of Euler angles or use a Euclidean distance between orientation representations, which again ignores Riemmanian properties like geodesic distances on the space of orientations. Another issue with regression-based methods is that they predict a single orientation without a confidence score. This is a problem for object categories 
%like boat and dining-table 
that exhibit strong symmetries because the regression system will predict a single orientation which might be far from the ground truth. In such cases, it would be preferable to have a system that predicts multimodal orientation distributions with a confidence score for each hypothesis.
Finally, another disadvantage of existing methods is that network architectures and training methods are not specifically designed for orientation estimation. For example, existing data augmentation techniques such as 2D jittering are adequate for object classification and detection, but some form of 3D jittering is needed for orientation estimation.

\subsection{Paper Contributions}

This work addresses the aforementioned drawbacks by designing a new 3D orientation estimation system that incorporates the geometry of the space of rotations into a combined regression and classification framework.

Our first contribution is the use of orientation representations like axis-angle and quaternion as outputs of a CNN and geodesic distances on the space of rotations as loss functions of our CNN. While these representations and distances have been used extensively in the past in computer vision, see e.g., \cite{MASKS03}, it is only recently that they have been used in the context of CNNs, e.g., \cite{Kendall:ICCV15} use the quaternion representation with an Euclidean loss function. 

Our second contribution is a framework for combining regression and classification-based approaches to orientation estimation. As we discussed earlier, pure regression and classification-based approaches both have certain disadvantages. However, by combining both formulations, we can get the best of both worlds. Specifically, we propose a family of \textit{Bin \& Delta} models, where a Bin network predicts a discrete orientation by solving a classification task and a Delta network predicts a continuous refinement of the discrete orientation by solving a regression task. The proposed framework is flexible and allows for many choices of regression and/or classification losses, what the output of each network represents and how to combine them. While recent work does combine pose regression and classification methods (see \S\ref{sec:related_work} for details), such methods are particular cases of our more general framework, which is designed taking the geometry of the space of rotations into account.

Our third contribution is a 3D pose jittering strategy for the task of orientation estimation. 2D jittering involves translating the bounding box around an object in the image plane, which is adequate for 2D object detection and classification, but limited for 3D orientation estimation. In this work, we jitter the object in 3D using small rotations (both in-plane rotations via deviations of camera-tilt and out-of-plane rotations via deviations of azimuth and elevation angles). This is more appropriate to train orientation estimation networks and better reflects the geometry of the jittered augmented data.

Our fourth contribution is to obtain state-of-the-art performance on the challenging Pascal3D+ dataset under a variety of metrics using both ground-truth bounding boxes and bounding boxes returned by object detection systems. Specifically, we introduce 14 distinct Bin \& Delta models with different decision choices and compare their performance. We also present a detailed evaluation of various decision choices via an extensive ablation analysis. This is an empirical validation of our design choices and the advantages of using a geometrically-aware CNN system for orientation estimation.

To summarize, the contributions of our work are:
\begin{itemize}
\item The integration of Riemannian representations of rotations and geodesic loss functions for 3D orientation estimation into convolutional neural networks;
\item The integration of regression and classification approaches to predict continuous 3D orientations while modeling multimodal orientation distributions;
\item A new geometry-based data augmentation strategy that is appropriate for orientation estimation; and
\item An experimental evaluation of our models showing that they achieve state-of-the-art performance on the challenging Pascal3D+ dataset.
\end{itemize}

This paper is an extended version of our preliminary work \citep{Mahendran:ICCVW17,Mahendran:arxiv18} with new models and architectures, new activation and loss functions, and new experimental results. Specifically, we introduce the Riemannian Bin \& Delta models in \S\ref{sec:rbd} that improve upon the Log-Euclidean Bin \& Delta models in \cite{Mahendran:arxiv18} (\S\ref{sec:lebd} here). We also recognize two different relaxations in the Probabilistic Bin \& Delta models in \cite{Mahendran:arxiv18} and correspondingly split them into our Probabilistic Bin \& Delta (\S\ref{sec:pbd}), RelaXed Bin \& Delta (\S\ref{sec:xbd}) and RelaXed Probabilistic Bin \& Delta models (\S\ref{sec:xpbd}). We also expand our ablation analysis (\S\ref{sec:ablation}) relative to \cite{Mahendran:ICCVW17} with new results from our Geodesic Bin \& Delta models. Finally, we also expand results on orientation estimation with detected bounding boxes (\S\ref{sec:results_detected}) relative to \cite{Mahendran:ICCVW17} with new models and a more detailed evaluation with different object detection systems.

\begin{table*}
	\centering
	\begin{tabular}{|p{3.7cm}|p{.129\textwidth}|p{.05\textwidth}|p{.51\textwidth}|}
		\hline
		Name & Models & Section & Salient Features \\
		\hline
		Geodesic Regression & $\R_G$ & \S\ref{sec:geodesic_regression} & Geodesic regression loss on 3D pose outputs \\
		Euclidean Regression & $\R_E$ & \S\ref{sec:euclidean_regression} & Euclidean regression loss on 3D pose outputs \\
		Classification & $\mathcal{C}$ & \S\ref{sec:classification} & Classification on the discretized 3D pose space \\
		Geodesic Bin \& Delta & $\M_G$ \& $\M_G+$ & \S\ref{sec:gbd} & Bin \& Delta model with geodesic loss on the 3D pose outputs and cross-entropy loss on the discretized pose space \\
		Riemannian Bin \& Delta & $\M_R$ \& $\M_R+$ & \S\ref{sec:rbd} & A Geodesic Bin \& Delta model with a different (Riemannian) choice of Delta representation \\
		Probabilistic Bin \& Delta & $\M_P$ \& $\M_P+$ & \S\ref{sec:pbd} & A Geodesic Bin \& Delta model with a probabilistic weighting of the geodesic loss \\
		RelaXed Bin \& Delta & $\M_X$ \& $\M_X+$ & \S\ref{sec:xbd} & A Geodesic Bin \& Delta with a soft-assignment to pose labels and KL-divergence loss \\
		RelaXed Probabilistic Bin \& Delta & $\M_{XP}$ \& $\M_{XP}+$ & \S\ref{sec:xpbd} & A Geodesic Bin \& Delta model with both probabilistic weighting of geodesic loss and soft-assignment to pose-labels \\
		Simple Bin \& Delta & $\M_S$ \& $\M_S+$ & \S\ref{sec:sbd} & Bin \& Delta model with Euclidean regression loss on the delta outputs and cross-entropy loss on the discretized pose space \\
		Log-Euclidean Bin \& Delta & $\M_{LE}$ \& $\M_{LE}+$ & \S\ref{sec:lebd} & Bin \& Delta model with Log-Euclidean regression loss on the delta outputs and cross-entropy loss on the discretized pose space \\
		\hline
	\end{tabular}
\caption{An overview of our proposed models for monocular 3D pose estimation.}
\label{table:models_overview}
\end{table*}

\subsection{Paper Organization}
The remainder of the paper is organized as follows. In \S\ref{sec:related_work}, we review the state of the art on 3D pose estimation. In \S\ref{sec:geometry}, we review some geometric properties of the space of orientations that we would like to use in designing our networks for orientation estimation, namely, representations of orientation in \S\ref{sec:representations} and geodesic loss functions in \S\ref{sec:geodesic_loss}. Then in \S\ref{sec:pure}, we describe our proposed geodesic regression network for orientation estimation that uses these representations and loss functions. In \S\ref{sec:mixed}, we discuss our proposed Bin \& Delta models (outlined in Table~\ref{table:models_overview}) that combine classification and regression for orientation estimation. This is followed by \S\ref{sec:data_augmentation} where we explain our geometry-based data augmentation strategy. In \S\ref{sec:results}, we demonstrate the effectiveness of our models with state-of-the-art performance on the Pascal3D+ dataset and a detailed experimental evaluation that includes results with bounding boxes returned by an oracle \S\ref{sec:3d_pose_results} and an object detection system \S\ref{sec:results_detected}. Finally, in \S\ref{sec:conclusion}, we state conclusions derived from our work.

\section{Related Work}
\label{sec:related_work}

\myparagraph{3D Pose Estimation Algorithms} There are many non-deep learning methods for 3D pose estimation given 2D images  \citep{Sastre:ICCVW11,Hejrati:NIPS12,Hejrati:CVPR14,Aubry:CVPR14,Glasner:ICCV11,Liebelt:CVPR10,Lim:ICCV13,Lim:ECCV14,Pepik:CVPR12,Pepik:ECCV12,Savarese:ICCV07,Savarese:ECCV08}. However, due to space constraints, we restrict our review to methods based only on deep networks. As mentioned earlier, the current literature on 3D pose estimation using deep networks can be divided in two groups: (i) methods that predict 2D keypoints from images and then recover 3D pose by solving a PnP problem, and (ii) methods that directly predict 3D pose from an image. 

The first group of methods includes the works of \cite{Crivellaro:ICCV15,Grabner:CVPR18,Pavlakos:ICRA17,Wu:ECCV16,Rad:ICCV17}. \cite{Pavlakos:ICRA17} and \cite{Wu:ECCV16} use 2D keypoints corresponding to the projections of semantically meaningful points on 3D object models as the output of their network. Given a new image, they predict a probabilistic map of 2D keypoints and recover 3D pose by comparing with some pre-defined object models. In \cite{Grabner:CVPR18,Rad:ICCV17,Crivellaro:ICCV15}, instead of semantic keypoints, they use 2D keypoints corresponding to the projection of the 8 corners of a 3D bounding box encapsulating the object. The network is trained by comparing the predicted 2D keypoint locations with the projections of the 3D keypoints on the image under ground-truth pose annotations. \cite{Grabner:CVPR18} uses a Huber loss on the projection error to be robust to inaccurate ground-truth annotations and is the current state-of-the-art on the Pascal3D+ dataset to the best of our knowledge.

The second group of methods includes the works of \cite{Tulsiani:CVPR15,Su:ICCV15,Elhoseiny:ICML16,Massa:BMVC16,Massa:arxiv14,Wang:PCM16,Mousavian:CVPR17}. All these methods train a network to predict the orientation given 2D images and use the Euler angle representation of rotation matrices to estimate the azimuth, elevation and camera-tilt angles separately. \cite{Tulsiani:CVPR15} and \cite{Elhoseiny:ICML16} divide the angles into non-overlapping bins and solve a classification problem. \cite{Su:ICCV15} additionally use a weighted cross-entropy loss for fine-grained orientation classification. On the other hand, \cite{Wang:PCM16} regresses the angles directly with a direct Euclidean loss. \cite{Massa:BMVC16} proposes multiple loss functions based on regression and classification for joint object detection and pose estimation, and concludes that the classification methods work better than the regression ones. \cite{Mousavian:CVPR17} divides the angles into overlapping bins and trains their network using a multibin loss that combines regression and classification. 
As we discussed earlier, our work is closest to the second ground of methods in that we are also interested in a monocular orientation estimation problem, but with the key difference that we are interested in designing our orientation estimation system with representations, loss functions and data augmentation that is geometrically coherent.

\myparagraph{3D Pose Representations and Loss Functions} There has been recent interest in quaternion representations of 3D pose for camera localization \citep{Kendall:ICCV15,Kendall:ICRA16,Kendall:CVPR17,Tulsiani:CVPR17}, but these works recommend a Euclidean, reprojection or classification loss respectively instead of the geodesic loss. Another work \citep{Wang:arxiv18}, uses quaternion representation (with a smoothed $L_1$ loss) in combination with a 3D location field to predict 3D pose of cars within a joint object detection and pose estimation framework. Recently, \cite{Hou:arxiv18} also propose Riemannian loss and gradients on the $SE(3)$ manifold in the context of CNNs for the task of pose estimation for image registration of fetal MRIs.

Our proposed mixed classification-regression framework, with its \emph{Bin and Delta models}, can be considered a generalization of \citep{Mousavian:CVPR17,Li:arxiv18,Guler:CVPR17,Guler:arxiv18}. Specifically, \cite{Mousavian:CVPR17} is a variation of the \emph{Geodesic Bin and Delta model} we propose in Eqn.~\eqref{eqn:geodesicbd} with a $\cos-\sin$ representation of Euler angles, while \cite{Li:arxiv18} is a particular case of the \emph{Simple Bin and Delta model} we propose in Eqn.~\eqref{eqn:m0+} with a quaternion representation of 3D pose. On the other hand, the quantized regression model of \cite{Guler:CVPR17,Guler:arxiv18} uses the Bin and Delta model to generate dense correspondences between a 3D model and an image for face landmark and human pose estimation. \cite{Guler:CVPR17} learns a modification of our \emph{Simple Bin and Delta model} in Eqn.~\eqref{eqn:m0} with a separate Delta network for every facial region while \cite{Guler:arxiv18} is a particular case of the \emph{Probabilistic Bin and Delta model} we propose in Eqn.~\eqref{eqn:probabilisticbd}. \cite{Guler:arxiv18} also makes a connection between a Bin and Delta model and a mixture of regression experts proposed in \cite{Jordan:NC94}, where the classification output probability vector acts as a gating function on regression experts.
There has been recent interest in trying to design networks and representations that combine classification and regression to model 3D pose but the authors of these different works have treated this as a one-off representation problem. In contrast, we propose a general framework that encapsulates prior models as particular cases.

\section{Preliminaries on 3D Orientations}
\label{sec:geometry}

In this section, we provide a brief overview of representations and loss functions of orientations that respect the special geometry of the space of orientations. We direct the reader to \cite{MASKS03} and \cite{Hartley-Zisserman04} for more details.

\subsection{Representations of Orientation}
\label{sec:representations}

We are interested in estimating the orientation, represented by a 3D rotation $R$, of an object in the image. There are multiple ways in which 3D rotations can be represented and we describe three of them here: (1) Euler angles, (2) Axis-angles and (3) Quaternions. We use the axis-angle representation in this work as it best captures the geometry and complexity of the space of 3D orientations.

\myparagraph{Euler Angles} A common way of describing rotation matrices is in terms of its Euler angles: azimuth $az$, elevation $el$ and camera-tilt $ct$. Assuming a ZXZ convention, the rotation matrix $R$ is given by
\begin{equation}
R = \begin{pmatrix} cc & -sc & 0 \\ sc & cc & 0 \\ 0 & 0 & 1 \end{pmatrix} \begin{pmatrix} 1 & 0 & 0 \\ 0 & ce & -se \\ 0 & se & ce \end{pmatrix} \begin{pmatrix} ca & -sa & 0 \\ sa & ca & 0 \\ 0 & 0 & 1 \end{pmatrix},
\end{equation}
where $ca=\cos (az)$, $sa=\sin (az)$, $ce = \cos (el)$, $se = \sin (el)$, $cc = \cos (ct)$, $sc = \sin (ct)$ and the three matrices capture rotations about the $Z$-axis by angle $ct$, about the $X$-axis by angle $el$ and about the $Z$-axis by angle $az$ respectively. However, this representation of 3D rotations, $(az, el, ct)$, suffers from the following two drawbacks: (1) The representation is not unique when $el=0$, a problem also commonly known as the Gimbal Lock. (2) The representation ignores properties of the space rotation matrices. Specifically, any rotation matrix $R$ lies in the set of special orthogonal matrices $SO(3) \doteq \{ R: R \in \mathbb{R}^{3 \times 3}, R^T R = I_3, \det(R) = 1 \}$. This space is a Lie group with an underlying Riemannian geometry, geodesic distance, etc. which are ignored in the per-angle representations and loss functions. Instead, we propose the use of two representations, axis-angle and quaternion, that better capture these properties.

\myparagraph{Axis-angle} A rotation matrix $R$ captures the rotation of 3D points by an angle $\theta \in [-\pi,\pi)$ about a unit-norm axis $v\in\mathbb{R}^3$. This can be expressed as $R = \expm(\theta [v]_{\times})$, where $\expm$ is the matrix exponential and $[v]_\times$ is the skew-symmetric operator of vector $v$, \ie $[v]_\times = [[0, -v_3, v_2]$, $[v_3, 0, -v_1]$, $[-v_2, v_1, 0]]$ for $v = [v_1, v_2, v_3]^T$ such that $\|v\|_2=1$. This can also be expressed as the exponential map $\exp : \mathbb{R}^3 \rightarrow SO(3)$ between axis-angle vectors $y \in \mathbb{R}^3$ and rotations $R \in SO(3)$.  Similarly, we can also define a logarithm map between a rotation $R$ and its axis-angle vector $y=\theta v$, $\log: SO(3) \rightarrow \mathbb{R}^3$. Restrict $\theta \in [0, \pi)$ and defining $R=I_3 \Leftrightarrow y=0_3$ ensures an invertible mapping. 

\myparagraph{Quaternion} Another popular representation for 3D rotation matrices are 
quaternions. Given an axis-angle vector $y = \theta v$, the corresponding quaternion $q = (c, s)$ is given by $(\cos \frac{\theta}{2}, \sin \frac{\theta}{2} v)^T$. By construction, quaternions are unit norm, \ie $\|q\|_2=1$, and are points on the hypersphere $S^3 \doteq \{ x: x \in \mathbb{R}^4, \|x\|_2 = 1 \}$. Also, note that quaternions $q$ and $-q$, or antipodal points on the hypersphere $S^3$, correspond to rotations by angle $\theta$ about axis $v$ and angle $2\pi-\theta$ about axis $-v$ respectively, which correspond to the same rotation matrix. Axis-angle vectors represent 3-degree of freedom rotations with a compact 3-dimensional representation whereas quaternions use a 4-dimensional representation with an additional unit-norm constraint. 

\subsection{Loss functions on 3D Orientation space}
\label{sec:geodesic_loss}

As we mentioned earlier, we would like to exploit the special properties of rotation matrices. One such property is the geodesic distance or the shortest distance between two points along the manifold, which is a better measure of the ``closeness'' between two elements of the manifold when compared to the Euclidean distance. The geodesic distance between two rotation matrices $R_1$ and $R_2$ is given by 
\begin{equation}
d(R_1, R_2) = \frac{\|\operatorname{logm}(R_1 R_2^T) \|_F}{\sqrt{2}}, 
\label{eqn:geodesicR}
\end{equation}
where $\operatorname{logm}$ is the matrix logarithm and $\|\cdot\|_F$ is the Frobenius norm. This can be simplified further using the Rodrigues' rotation formula, 
\begin{equation}
R = I_3 + \sin \theta [v]_\times + (1 - \cos \theta) [v]_\times^2, 
\label{eqn:Rodrigues}
\end{equation}
to get 
\begin{equation}
d(R_1, R_2) = | \cos^{-1} \left[ \frac{tr(R_1^T R_2) - 1}{2} \right] | .
\label{eqn:distR}
\end{equation}
This geodesic distance better captures the (Riemannian) geometry of the orientation space and is the loss function we would like to use for our proposed work. 

\section{Geodesic Regression Network for Orientation Estimation}
\label{sec:pure}

In this section, we present our proposed network that formulates the orientation estimation task as a geodesic regression problem. We present the network architecture in \S\ref{sec:architecture_pure}, loss function in \S\ref{sec:loss_pure} and model in \S\ref{sec:geodesic_regression}. We also present two baselines that formulate the task as Euclidean regression (\S\ref{sec:euclidean_regression}) or classification (\S\ref{sec:classification}). 

\subsection{Architecture}
\label{sec:architecture_pure}

\begin{figure}
	\includegraphics[width=\linewidth]{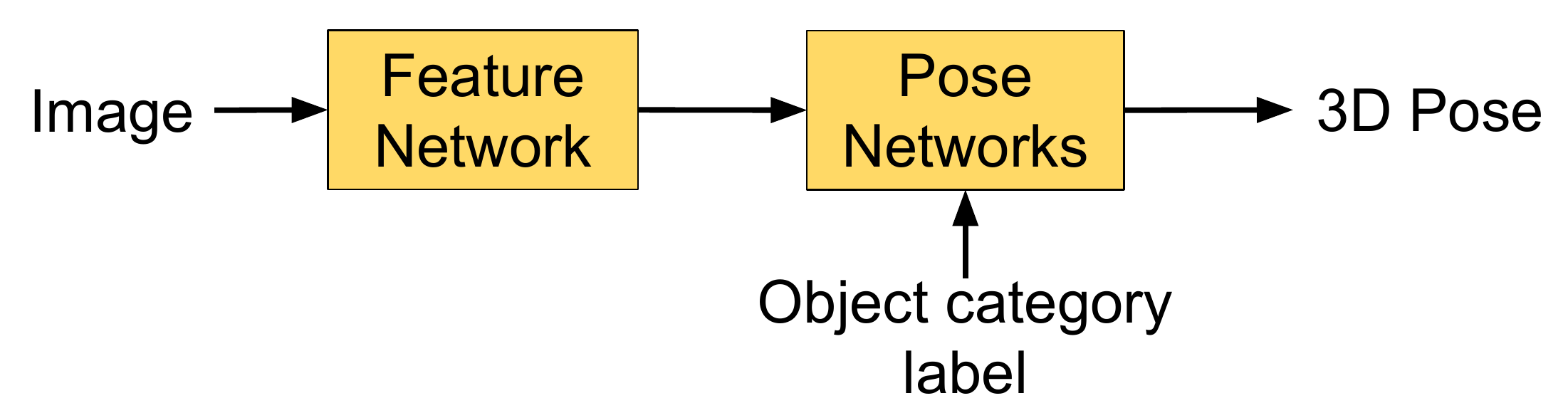} 
	\caption{Overall network architecture which takes image $x$ and object category label $c$ as input and predicts object pose $y$. The feature network $\Phi_F(\cdot; W_F)$ is shared across all object categories, while each category has its own pose network $\Phi_P(\cdot; W_P^c)$.}
	\label{fig:network_architecture}
\end{figure}

We follow a standard network design for 3D pose estimation, where the network has two parts: a feature network and a pose network, as illustrated in Fig.~\ref{fig:network_architecture}. The feature network is shared across all object categories, but there is one pose network associated with each object category.The feature network takes as input an Image $x$, and outputs the feature descriptor of the image, $\Phi_F(x; W_F)$, where $W_F$ are the weights associated with the feature network. Our feature network is the ResNet-50 network of \cite{He:CVPR16,He:ECCV16} minus the last classification layer. The pose network takes the object category label $c$ and the output of the feature network $f=\Phi_F(x; W_F)$ as input and predicts the object pose $y = \Phi_P(f; W_P^c)$, where $W_P^c$ denotes weights of the pose network associated with category label $c$.

\begin{figure}
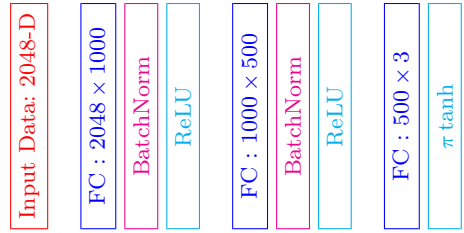

	\small
	\centering
	{\color{red}\rotatebox{90}{\framebox[3cm][c]{Input Data: $2048$-D}}}
	\quad
	{\color{blue}\rotatebox{90}{\framebox[3cm][c]{FC : $2048 \times 1000$}}}
	{\color{magenta}\rotatebox{90}{\framebox[3cm][c]{BatchNorm}}}
	{\color{cyan}\rotatebox{90}{\framebox[3cm][c]{ReLU}}}
	\quad
	{\color{blue}\rotatebox{90}{\framebox[3cm][c]{FC : $1000 \times 500$}}}
	{\color{magenta}\rotatebox{90}{\framebox[3cm][c]{BatchNorm}}}
	{\color{cyan}\rotatebox{90}{\framebox[3cm][c]{ReLU}}}
	\quad
	{\color{blue}\rotatebox{90}{\framebox[3cm][c]{FC : $500 \times 3$}}}
	{\color{cyan}\rotatebox{90}{\framebox[3cm][c]{$\pi \tanh$}}}
	\caption{A Pose Network for the axis-angle representation}
	\label{fig:pose_network}
\end{figure}

\begin{figure}
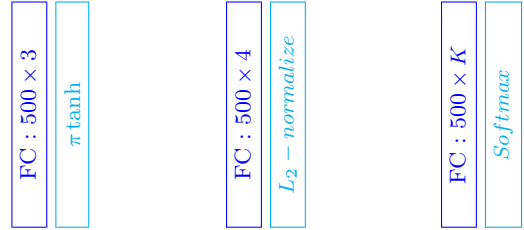

	\small
	\centering
	\begin{subfigure}[t]{0.3\linewidth}
		\centering
		{\color{blue}\rotatebox{90}{\framebox[3cm][c]{FC : $500 \times 3$}}}
		{\color{cyan}\rotatebox{90}{\framebox[3cm][c]{$\pi \tanh$}}}
		\caption{Axis-angle + regression}
		\label{fig:pose_network_axisangle}
	\end{subfigure}
	~
	\begin{subfigure}[t]{0.3\linewidth}
		\centering
		{\color{blue}\rotatebox{90}{\framebox[3cm][c]{FC : $500 \times 4$}}}
		{\color{cyan}\rotatebox{90}{\framebox[3cm][c]{$L_2-normalize$}}}
		\caption{Quaternion + regression}
		\label{fig:pose_network_quaternion}
	\end{subfigure}
	~
	\begin{subfigure}[t]{0.3\linewidth}
		\centering
		{\color{blue}\rotatebox{90}{\framebox[3cm][c]{FC : $500 \times K$}}}
		{\color{cyan}\rotatebox{90}{\framebox[3cm][c]{$Softmax$}}}
		\caption{Classification}
		\label{fig:pose_network_classification}
	\end{subfigure}
	\caption{Last 2 layers of the pose network architecture for different representations and loss functions}
	\label{fig:pose_network_differences}
\end{figure}

The pose network has 3 fully connected layers with associated ReLU activations and batch normalization as outlined in Fig.~\ref{fig:pose_network} for the axis-angle representation. The difference between different pose networks for axis-angle and quaternion representations is in the last 2 layers as shown in Fig.~\ref{fig:pose_network_differences}.
For the axis-angle representation, the output of the pose network is a 3-dim output $y=\theta v$ and we model the constraints $\theta \in [0, \pi)$ and $v_i \in [-1, 1]$ using a $\pi \tanh$ non-linearity.
For the quaternion representation, the output is 4-dimensional and we model the unit-norm constraint using a $L_2$ normalization non-linearity.

\subsection{Loss Functions}
\label{sec:loss_pure}

We would like to use the geodesic regression loss discussed in \S\ref{sec:geodesic_loss} as our loss function for Geodesic Regression. Under the two representations of axis-angle and quaternion, this loss function takes different forms. 

\myparagraph{Axis-angle Loss} The geodesic regression loss between two axis-angle vectors $y_1$ and $y_2$ is given by 
\begin{equation}
\mathcal{L}_R(y_1, y_2) \equiv d(R_1, R_2),
\label{eqn:geodesicAA}
\end{equation}
where $R_1=\exp(y_1)$ and $R_2=\exp(y_2)$ and $d(R_1, R_2)$ is defined in Eqn.~\eqref{eqn:distR}. This geodesic regression loss is our proposed alternative to the standard Euclidean regression loss, $\mathcal{L}_E(y_1, y_2) = \| y_1 - y_2 \|_2^2$. 

Note that another interpretation of axis-angle vectors $y = \log(R)$ is that they are projections of rotation matrices $R$ to the tangent space at the identity $I_3$. Under this interpretation of axis-angle vectors, the Euclidean regression loss $\mathcal{L}_E(y_1, y_2)$ is (locally) the Euclidean distance between projections of corresponding rotation matrices $R_1$ and $R_2$ in the tangent space at $I_3$. The Euclidean distance on the tangent space is also a popular choice of distances in Riemannian geometry but, this is a good distance only locally (near $I_3$) whereas the geodesic distance is valid globally. 

It is interesting to note that,
\begin{align}
\mathcal{L}_R(y_1, y_2) &= d(R_1, R_2) = \frac{\|\operatorname{logm}(R_1 R_2^T)\|_F}{\sqrt{2}} \nonumber \\
	&= \frac{1}{\sqrt{2}} \| \operatorname{logm} \big [ \expm(\theta_1 [v_1]_\times) \expm(\theta_2 [v_2]_\times)^T \big ] \|_F \nonumber \\
	&\neq \frac{1}{\sqrt{2}} \| \operatorname{logm} \big [ \expm(\theta_1 [v_1]_\times-\theta_2 [v_2]_\times) \big ] \|_F \nonumber \\
	&= \frac{1}{\sqrt{2}} \| \theta_1 [v_1]_\times - \theta_2 [v_2]_\times \|_F = \|\theta_1 v_1 - \theta_2 v_2 \|_2 \nonumber \\
	&= \|y_1 - y_2 \|_2 = \mathcal{L}_E(y_1, y_2)
\end{align}

This is another way of saying that the Euclidean distance between two axis-angle vectors is not a geometrically appropriate loss function. 

\myparagraph{Quaternion Loss} The geodesic distance between two quaternions $q_1$ and $q_2$ is given by 
\begin{equation}
d(q_1, q_2) = 2 \cos^{-1}(|c|), \hspace{1em} \text{where} \hspace{1em} (c, s) = q_1^{-1} \cdot q_2. 
\label{eqn:distQ}
\end{equation}
This is identical to the geodesic distance $d(R_1, R_2)$ of Eqn.\eqref{eqn:geodesicR}, where $q_1$ and $q_2$ are quaternions corresponding to rotation matrices $R_1$ and $R_2$ respectively. Using quaternion algebra, we have $(c_1, s_1) \cdot (c_2, s_2) = (c_1c_2- \langle s_1, s_2 \rangle, c_1 s_2+c_2 s_1+s_1 \times s_2)$ and $(c, s)^{-1} = (c, -s)$ for unit norm $q=(c, s)$. Substituting these in Eqn.~\eqref{eqn:geodesicR} leads to the geodesic quaternion regression loss, 
\begin{equation}
\mathcal{L}_R(q_1, q_2) = 2 \cos^{-1}(| \langle q_1, q_2 \rangle|) .
\label{eqn:geodesicQ}
\end{equation}
 
\subsection{Geodesic Regression Model}
\label{sec:geodesic_regression}

We combine the network architecture and loss functions described earlier to form the Geodesic Regression network, which is trained while solving the optimization problem, 
\begin{equation}
\R_G:  \min_{W} \frac{1}{N} \sum_n \mathcal{L}_R(y_n^*, y_n).
\label{eqn:r1}
\end{equation}
Here $\mathcal{L}_R(y_n^*, y_n)$ is the geodesic loss (defined in Eqns.~\eqref{eqn:geodesicAA} and \eqref{eqn:geodesicQ}) between ground-truth pose $y_n^*$ and predicted pose $y_n = \Phi_P^{\R}(\Phi_F(x_n; W_F); W_P^c)$ returned by the network given input image $x_n$, and $W = [W_F, \{W_P^c\}]$ includes all the parameters of the feature network and per-category regression pose networks. 

\subsection{Baselines}
\label{sec:baselines}

We compare the Geodesic Regression network with two common-sense baselines of (i) Euclidean Regression and (ii) Classification, which we describe next. 

\subsubsection{Euclidean Regression Model}
\label{sec:euclidean_regression}

In this model, we train the network using the squared Euclidean loss instead of the Geodesic loss leading to the optimization problem,
\begin{equation}
\R_E: \min_{W} \frac{1}{N} \sum_n \| y_n^* - y_n \|_2^2.
\label{eqn:r0}
\end{equation}
As we shall see later in \S\ref{sec:3d_pose_results} (Table~\ref{table:regression}) and \S\ref{sec:ablation} (Table~\ref{table:quaternion_results}), the Geodesic regression model is consistently better than the Euclidean regression model for both axis-angle and quaternion representations as it models the underlying geometry of the problem more faithfully. 

\subsubsection{Classification Model}
\label{sec:classification}

Here, we discretize the continuous pose space into bins defined by a given dictionary of key poses $\{z_k\in\R^3\}_{k=1}^K$. Formally, given a pose $y\in\R^3$ (axis-angle representation), we assign it pose label $l\in\{1,\dots,K\}$ if key pose $z_l$ is the closest to $y$, i.e.,  $l = \operatorname{argmin}_k \|y - z_k\|_2$. The work of \cite{Li:arxiv18} used a pre-specified dictionary of key poses obtained by a uniform tessellation of $SO(3)$. In this work, we try to capture the geometry of the data (target orientation space) by learning a K-Means dictionary $\{z_k\}$ from training data as our key pose dictionary. Note that our pose target $y$ can be any choice of orientation representation: axis-angle or quaternion representations. Then, the network architecture is the same as for the regression models with the difference being the last 2 layers of the pose network (shown in Fig.~\ref{fig:pose_network_classification}). Notice that the output of the classification model is $K$-dimensional, where $K$ is the size of the pose dictionary used to discretize the pose space and the final non-linearity is a softmax operation. The classification model is then trained by solving the optimization problem,
\begin{equation}
\mathcal{C}: \min_W \frac{1}{N} \sum_n \mathcal{L}_c(l_n^*, l_n),
\label{eqn:c0}
\end{equation}
where $\mathcal{L}_C(l_n^*, l_n)$ is the cross-entropy loss between ground-truth pose label, 
\begin{equation}
l_n^* = \operatorname{argmin}_k \|y_n^* - z_k \|_2, 
\label{eqn:pose_label}
\end{equation}
and predicted pose label $l_n = \Phi_P^{\mathcal{C}}(\Phi_F(x_n; W_F); W_P^c)$. The final pose output is given by $y_n = z_{l_n}$. 

\section{Mixed Classification-Regression Networks for Orientation Estimation}
\label{sec:mixed}

As mentioned earlier, a disadvantage of the pure regression approach is that it is unable to properly model multimodal pose distributions arising in object categories that exhibit shape symmetries. One way to overcome this limitation is to break down the task of estimating 3D pose into three parts: (1) estimate a distribution of discrete pose labels associated with some key poses, a classification task that can capture the multimodal nature of the pose space, (2) estimate a continuous deviation from the key poses, a regression task that still returns fine pose estimates, and (3) combine the discrete and continuous pose estimates via some combination function. These three steps are achieved using our Bin \& Delta models, which we now describe in more detail. Specifically, \S\ref{sec:architecture_mixed} describes the proposed network architecture, \S\ref{sec:loss_mixed} describes the proposed loss functions, and  \S\ref{sec:gbd}-\S\ref{sec:lebd} describe a variety of models that arise from different choices of the combination function, regression loss, classification loss and network architecture.

\subsection{Architecture}
\label{sec:architecture_mixed}
\begin{figure}
	\includegraphics[width=\linewidth]{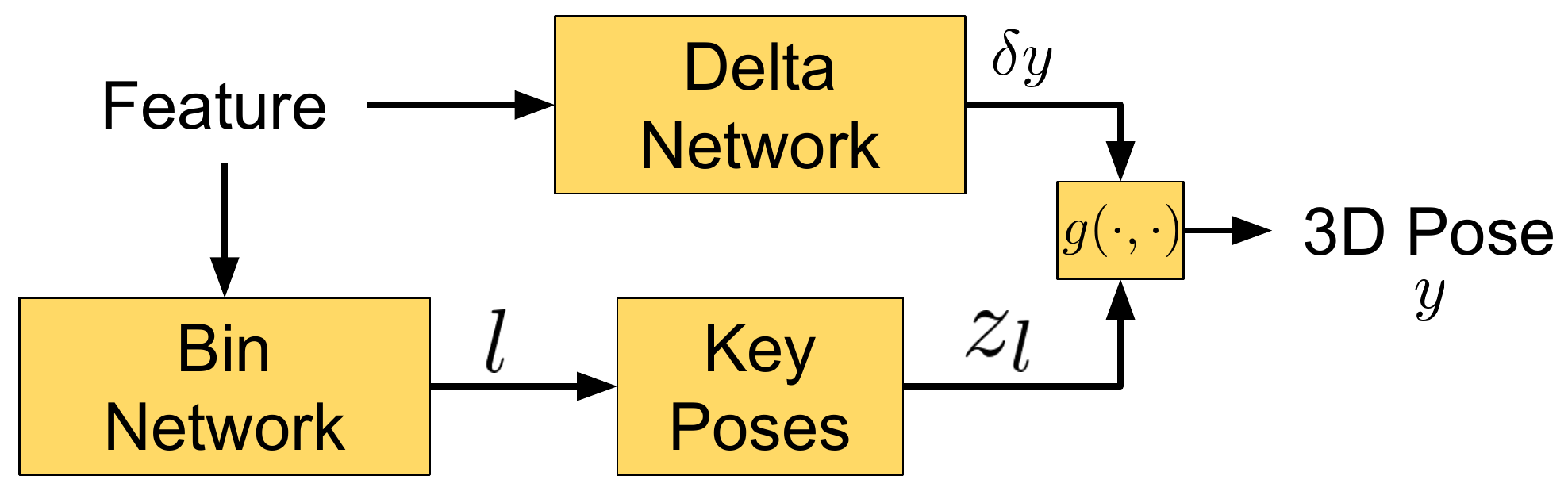}
	\caption{Network architecture of the Bin \& Delta model}
	\label{fig:bin_and_delta}
\end{figure}

Instead of using a single multi-layer perceptron as the pose network, as done in the pure regression and classification networks shown in \S\ref{sec:pure}, Fig.~\ref{fig:pose_network}, the Bin \& Delta model has two components: a Bin network (for classification) and a Delta network (for regression) as shown in Fig.~\ref{fig:bin_and_delta}. Both networks take as input the output of the feature network, $f$. The Bin network predicts a pose-label $l = \Phi_B(f; W_B^c)$,
%\footnote{rv: why this doesn't depend on object category $c$? sm: corrected} 
where $\Phi_B$ is the Bin network parameterized by weights $W_B^c$ associated with object category $c$ and the pose-label $l$ references a key pose $z_{l}$. The Delta network predicts a pose-residual $\delta y = \Phi_D(f; W_D^c)$ with Delta network $\Phi_D$ and its weights $W_D^c$. Given a classification output, pose label $l$, and a regression output $\delta y$ are combined as
\begin{equation}
y = g(z_l, \delta y),
\label{eqn:b&d}
\end{equation}
where $z_l$ is the key pose corresponding to label $l$, $\delta y$ is the deviation from that key pose and $g(\cdot, \cdot)$ is some combination function. Therefore, the parameters of the  Bin \& Delta  model are given by $W = [W_F, \{W_B^c\}, \{W_D^c\}]$.

\begin{figure}
	\includegraphics[width=\linewidth]{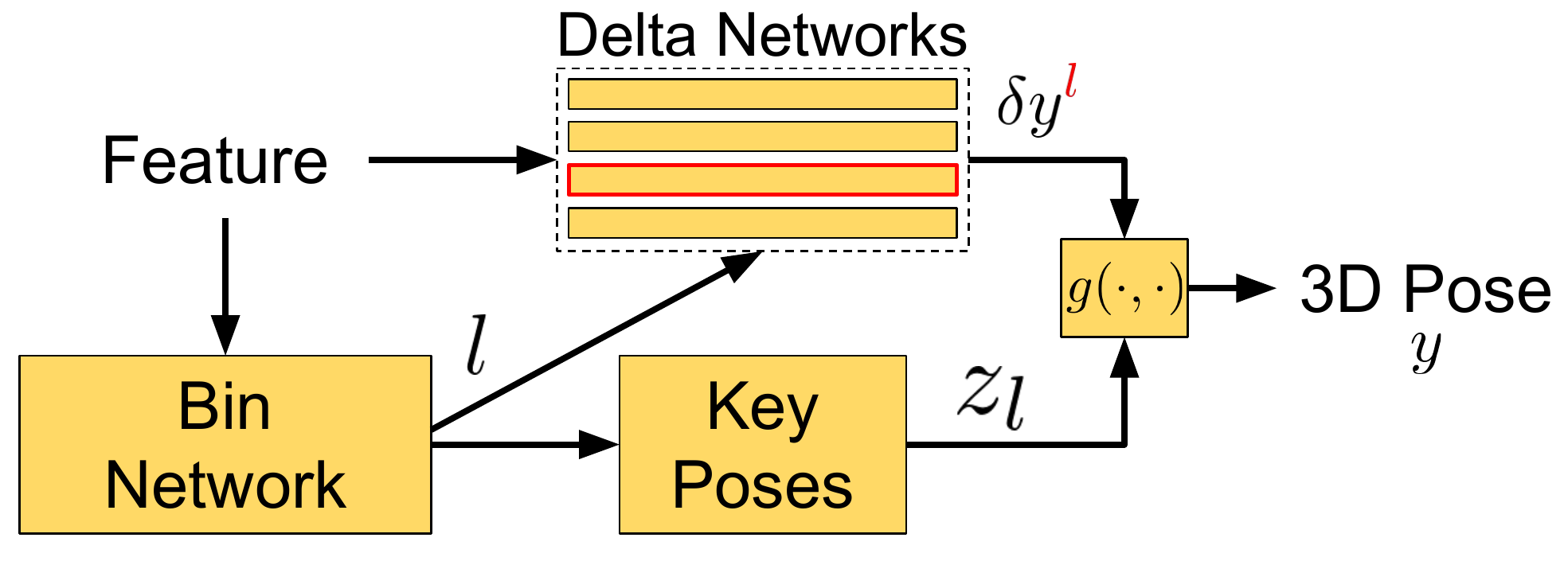}
	\caption{One Delta network per pose-bin}
	\label{fig:multiple_delta}
\end{figure}

When designing the combination function two major choices arise. One choice is to have a single pose-residual network for all key poses, in which case the value of $\delta y$ does not depend on the pose label $l$. An alternative choice shown in Fig.~\ref{fig:multiple_delta} is to have a Delta model for every single pose-bin in which case, $\delta y \rightarrow \delta y^l$ will also be a function of the pose label $l$. This modeling decision is equivalent to deciding whether to have a common covariance matrix across all clusters or have a different covariance matrix for every cluster in a Gaussian Mixture Model (GMM).

\subsection{Loss Functions}
\label{sec:loss_mixed}
As implied by the name, the loss function for a Mixed Classification-Regression network is a combination of a regression loss and a classification loss. Specifically, the loss is given by
\begin{equation}
\mathcal{L}(y_1, y_2) = \alpha \mathcal{L}_R(g(z_{l_1}, \delta y_1), g(z_{l_2}, \delta y_2)) + \mathcal{L}_C(l_1, l_2),
\label{eqn:mixed_loss}
\end{equation}
where we apply the geodesic regression loss $\mathcal{L}_R$ on the pose outputs $y_1 = g(z_{l_1}, \delta y_1)$ \& $y_2 = g(z_{l_2}, \delta y_2)$ and the classification loss $\mathcal{L}_C$ on pose labels $l_1$ \& $l_2$. The losses are combined with a relative weighting parameter $\alpha$. The pose variables $y_1$ and $y_2$ can refer to any choice of orientation representation with an appropriately chosen combination function. The mixed loss of Eqn.~\eqref{eqn:mixed_loss} is an extension of the geodesic loss of Eqns.~\eqref{eqn:geodesicAA}, \eqref{eqn:geodesicQ} 
for the mixed representation with additional supervision on the classification component. Note that applying the regression loss at the pose output is also a modeling choice and as we shall see later in \S\ref{sec:sbd} and \S\ref{sec:lebd}, we can instead apply a regression loss on the output of the Delta network to get a different model.

\subsection{Models}
\label{sec:models_mixed}

Combining the network architecture and loss functions described above with different modeling choices leads to a variety of Bin \& Delta models which are detailed in this section. An overview of all the models discussed here is provided in Table~\ref{table:models_overview}. 

\subsubsection{Geodesic Bin \& Delta Model}
\label{sec:gbd}

The Geodesic Bin \& Delta model is our first mixed classification-regression model, where we choose the combination function $g(\cdot, \cdot)$ in Eqn.~\eqref{eqn:b&d} as
\begin{equation}
g(z_l, \delta y) = z_l + \delta y.
\label{eqn:sum_combination}
\end{equation}
This choice of $g$ when substituted in the mixed-loss function of Eqn.~\eqref{eqn:mixed_loss}, leads to the optimization problem
\begin{equation}
\M_G: \min_W  \frac{1}{N} \sum_n \Big [ \alpha \mathcal{L}_R(y_n^*, z_{l_n} + \delta y_n) + \mathcal{L}_C(l_n^*, l_n) \Big ],
\label{eqn:geodesicbd}
\end{equation}
where we minimize the weighted sum of two terms: (1) Geodesic loss $\mathcal{L}_R$ between ground-truth pose $y_n^*$ and predicted pose $y_n = z_{l_n} + \delta y_n$, and (2)  Cross-entropy loss $\mathcal{L}_C$ between ground-truth pose label $l_n^*$ (given by Eqn.~\eqref{eqn:pose_label}) and predicted pose label $l_n$ returned by the Bin network. 

\begin{figure}
\centering
\includegraphics[width=0.5\linewidth]{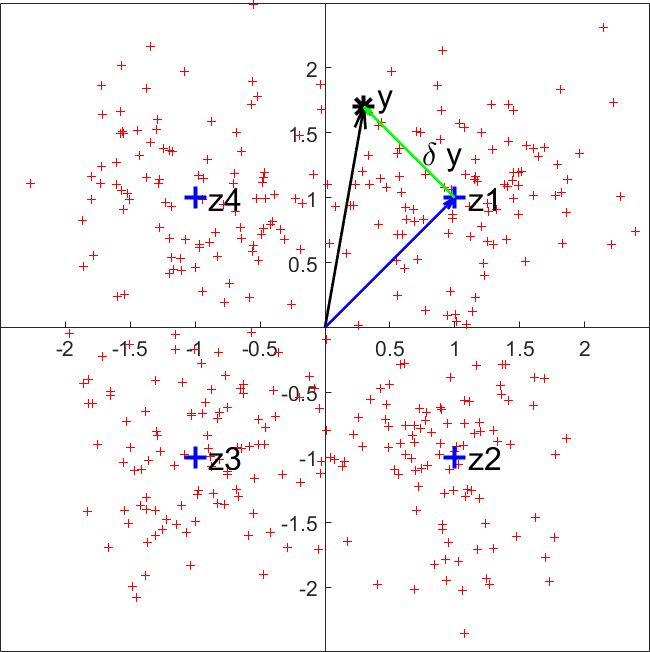}
\caption{Toy example of the Geodesic Bin \& Delta model for a plane. (Best seen in color)}
\label{fig:geodesic_bin_delta}
\end{figure}

A toy example of the Geodesic Bin \& Delta model is illustrated in Fig.~\ref{fig:geodesic_bin_delta} in the context of points on a plane. Given training data (the points shown in red),  K-Means clustering returns the dictionary of key poses: $\{z_1, z_2, z_3, z_4\}$ (shown in blue $+$). A new point $y$ (shown in a black $*$) with associated label $l=1$ is now a sum of key pose $z_1$ and residual $\delta y$ (shown in green). This is actually what is happening in the Geodesic Bin \& Delta model too, except that the X-Y plane is now the tangent plane on the rotation manifold at the identity $I_3$ and the points are 3-dimensional instead of the 2-D points shown in the toy example in Fig.~\ref{fig:geodesic_bin_delta}.

Using one Delta network per pose-bin (Fig.~\ref{fig:multiple_delta}) instead of only one Delta network for all pose-bins (Fig.~\ref{fig:bin_and_delta}) in the Geodesic Bin \& Delta model leads to the optimization problem
\begin{equation}
\M_G+: \min_W  \frac{1}{N} \sum_n \Big [ \alpha \mathcal{L}_R(y_n^*, z_{l_n} + \delta y_n^{\textcolor{red}{l_n}}) + \mathcal{L}_C(l_n^*, l_n) \Big ],
\label{eqn:geodesicbd+}
\end{equation}
where the modification is highlighted in red. This leads to more flexibility in modeling the pose residuals. 

Notice that everything we have discussed so far is valid for both axis-angle and quaternion representations, with the only modification for quaternions being 
\begin{equation}
g(z_l, \delta y) = \frac{z_l + \delta y}{\|z_l + \delta y\|_2},
\end{equation}
to ensure the unit-norm constraint.

\subsubsection{Riemannian Bin \& Delta Model}
\label{sec:rbd}

Another way to interpret the pose residual in the Geodesic Bin \& Delta model is
\begin{equation}
\delta y = y - z_l = \log(R) - \log(\tilde{R}_l),
\label{eqn:geodesic_delta}
\end{equation}
where $\tilde{R}_l = \exp(z_l)$ is the key rotation associated with key pose $z_l$. As we mentioned earlier, the $\log$ operation projects the rotation matrix $R$ onto the tangent plane at the identity, $I_3$, and Eqn.~\eqref{eqn:geodesic_delta} defines the pose residual as a difference in the tangent plane. We propose an alternative definition of the pose residual
\begin{equation}
\delta y = \log_{\tilde{R}_l}(R),
\label{eqn:riemannian_delta}
\end{equation}
where we now project the rotation matrix R onto the tangent plane at key rotation $\tilde{R}_l$. This is equivalent to defining the combination function $g(\cdot, \cdot)$ as
\begin{equation}
g(z_l, \delta y) = \log ( \exp(z_l) \exp(\delta y)).
\label{eqn:riemannian_delta2}
\end{equation}
We call this the Riemannian Bin \& Delta model because we use the Riemannian exponential and logarithm maps in defining what the output of the Bin and Delta networks represent. The pose output is now given by
\begin{equation}
R = \tilde{R}_l \exp(\delta y).
\label{eqn:riemannian_output}
\end{equation}

Substituting this into the mixed loss of Eqn.~\eqref{eqn:mixed_loss}, we get the optimization problem 
\begin{equation}
\M_R: \min_W  \frac{1}{N} \sum_n \Big [ \alpha \mathcal{L}_R(R_n^*, \tilde{R}_{l_n}\exp(\delta y_n)) + \mathcal{L}_C(l_n^*, l_n) \Big ],
\label{eqn:riemannianbd}
\end{equation}
where we again minimize the weighted sum of two terms: (1) Geodesic loss $\mathcal{L}_R$ between ground-truth rotations $R_n^*$ and predicted rotations $R_n$, and (2) Cross-entropy loss $\mathcal{L}_C$ between ground-truth class label $l_n^*$ and predicted class label $l_n$.

\begin{figure}
\centering
\includegraphics[width=0.5\linewidth]{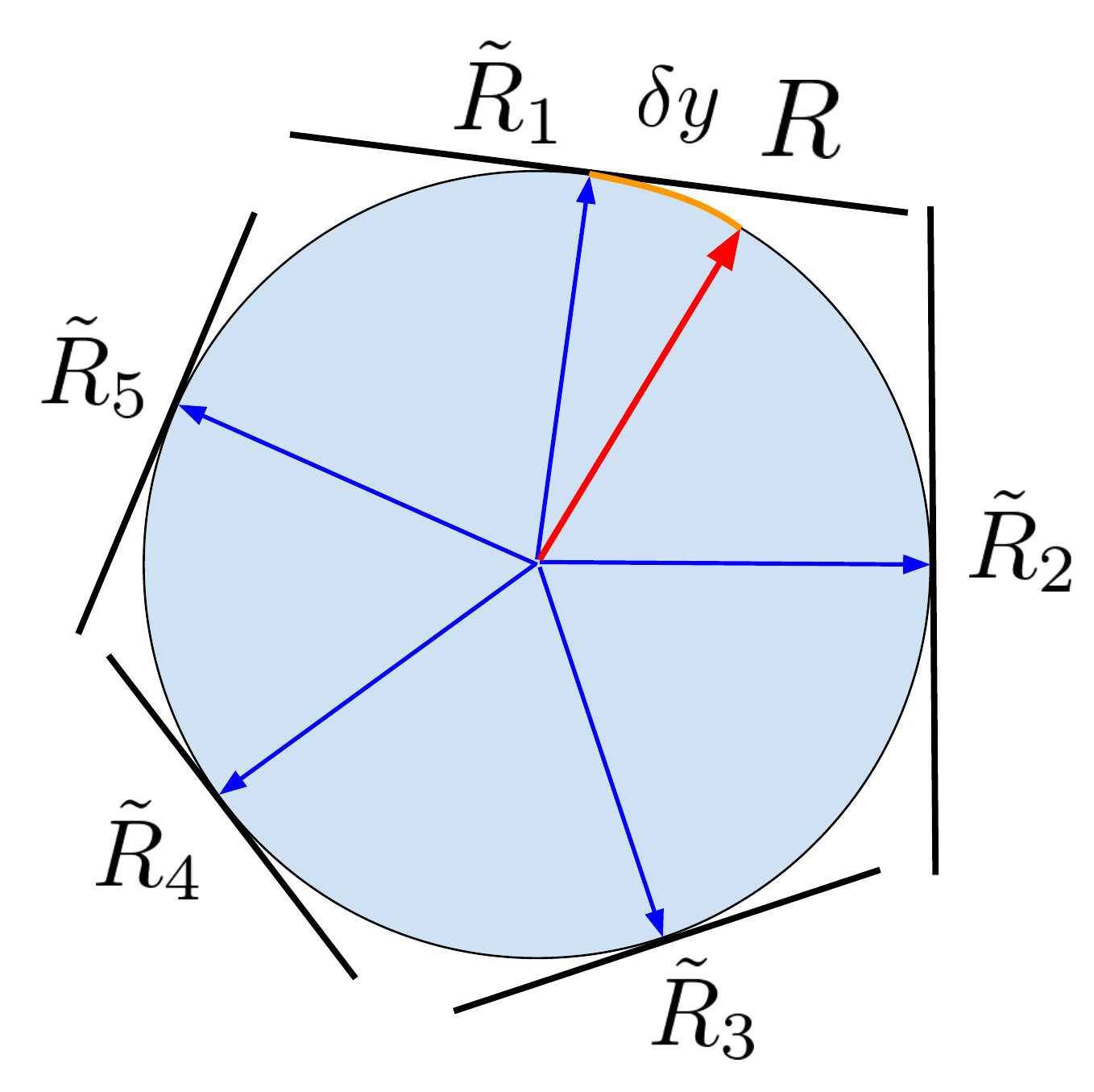}
\caption{Toy example of the Riemannian Bin \& Delta model for a circle. (Best seen in color).}
\label{fig:rotation_bin_delta}
\end{figure}

A toy example of the Riemannian Bin \& Delta model for a circle is shown in Fig.~\ref{fig:rotation_bin_delta}. The figure shows a circle with 5 tangent planes corresponding to key poses $\tilde{R}_i, i=1,...,5$. The rotation $R$ (shown in red) is now a combination of the key pose $\tilde{R}_1$, with pose label $l=1$, and the delta $\delta y$ (shown in orange).

Again, using one Delta network per pose-bin we get the modified optimization problem
\begin{equation}
\M_R+: \min_W  \frac{1}{N} \sum_n \Big [ \alpha \mathcal{L}_R(R_n^*, \tilde{R}_{l_n}\exp(\delta y_n^{\textcolor{red}{l_n}})) + \mathcal{L}_C(l_n^*, l_n) \Big ],
\label{eqn:riemannianbd+}
\end{equation}
where the difference is highlighted in red. Note that everything is defined in terms of rotations for the Riemannian Bin \& Delta model and we use only the axis-angle representation in this model, which are easily transformed to rotations via the Rodrigues' rotation formula. 

\subsubsection{Probabilistic Bin \& Delta Model}
\label{sec:pbd}

In the Geodesic Bin \& Delta model, given the output probabilities of the Bin network for image $x_n$, $p_{nk}, k=\{1, \hdots, K\}$, we predict the most likely pose label as $l_n = \operatorname{argmax}_k p_{nk}$ and use it during network training. A different formulation that better uses this probability output is our Probablistic Bin \& Delta model which solves the optimization problem
\begin{equation}
\M_P : \min_W  \frac{1}{N} \sum_n  \Big [ \alpha \sum_k p_{nk} \mathcal{L}_R(y_n^*, z_k + \delta y_n) + \mathcal{L}_C(l_n^*, l_n) \Big ],
\label{eqn:probabilisticbd}
\end{equation}
where we are now weighting the geodesic loss between the ground-truth pose $y_n^*$ and per-category predicted pose $y_n^k = z_k + \delta y_n$ with the the probability of the pose label being assigned that class, $p_{nk}$. The final predicted pose is still given by $y_n = z_{l_n} + \delta y_n$, where $l_n = \operatorname{argmax}_k p_{nk}$, but we expect to learn better models due to more information in the modified optimization problem. The one-delta per pose-bin version of this model is given by the optimization problem
\begin{align}
\begin{split}
\M_P+ : \min_W  \frac{1}{N} \sum_n \Big [ \alpha \sum_k p_{nk}  \mathcal{L}_R(y_n^*, z_k + \delta y_n^{\textcolor{red}{k}})& \\+ \mathcal{L}_C(l_n^*, l_n)& \Big ],
\label{eqn:probabilisticbd+}
\end{split}
\end{align}
with the difference highlighted in red again.

\subsubsection{RelaXed Bin \& Delta}
\label{sec:xbd}
Another relaxation in our original problem formulation is that instead of a hard assignment (via K-Means), where we assign a single key-pose to an image \ie $l_n^* = \operatorname{argmax}_k \|y_n^* - z_k\|_2$, a more flexible model and possibly more informative model would be to do a soft assignment to all key-poses and use this probabilistic information in a better way. Post K-Means, one can generate a probabilistic assignment using
\begin{equation}
p_{nk}^* = \frac{\exp(- \gamma \|y_n^* - z_k\|_2^2)}{\sum_k \exp(- \gamma \|y_n^* - z_k\|_2^2)}.
\label{eqn:soft}
\end{equation}
Now, the classification loss can be modified to be a Kullback-Leibler divergence $\mathcal{L}_{KD}$ between ground-truth and predicted probabilities. This modification in the Geodesic Bin \& Delta model of Eqn.~\eqref{eqn:geodesicbd} leads to the RelaXed Bin \& Delta model with the optimization 
\begin{equation}
\M_X : \min_W  \frac{1}{N} \sum_n  \Big [\alpha \mathcal{L}_R(y_n^*, z_{l_n} + \delta y_n) + \mathcal{L}_{KD}(p_n^*, p_n) \Big ],
\label{eqn:relaxedbd}
\end{equation}
where $p_n^* = [p_{nk}^*]_{k=1}^K$ and $p_{nk} = [p_{nk}]_{k=1}^K$ and its variant
\begin{equation}
\M_X+ : \min_W  \frac{1}{N} \sum_n \Big [ \alpha \mathcal{L}_R(y_n^*, z_{l_n} + \delta y_n^{\textcolor{red}{l_n}}) + \mathcal{L}_{KD}(p_n^*, p_n) \Big ].
\label{eqn:relaxedbd+}
\end{equation}

\subsubsection{RelaXed Probabilistic Bin \& Delta Model}
\label{sec:xpbd}
The same relaxation described in \S\ref{sec:xbd} can be applied to the Probabilistic Bin \& Delta model to get RelaXed-Probabilistic Bin \& Delta models:
\begin{align}
\M_{XP} : \min_W  \frac{1}{N} \sum_n  \Big [ & \alpha \sum_k p_{nk} \mathcal{L}_R(y_n^*, z_k + \delta y_n) \nonumber \\ & + \mathcal{L}_{KD}(p_n^*, p_n) \Big ], \label{eqn:relaxedprobbd} \\
\M_{XP}+ : \min_W  \frac{1}{N} \sum_n \Big [ & \alpha \sum_k p_{nk} \mathcal{L}_R(y_n^*, z_k + \delta y_n^{\textcolor{red}{k}}) \nonumber \\ & + \mathcal{L}_{KD}(p_n^*, p_n) \Big ]. \label{eqn:relaxedprobbd+}
\end{align}
These were the models proposed as Probabilistic Bin \& Delta previously \citep{Mahendran:arxiv18}
, but as can be seen above, they include two probabilistic changes and are studied separately as three different classes of models here. 

\subsubsection{Simple Bin \& Delta Model}
\label{sec:sbd}
Instead of applying the regression loss at the pose output, we can apply it at the output of the Delta network leading to our Simple Bin \& Delta model,
\begin{equation}
\M_S: \min_W \frac{1}{N} \sum_n \left [ \alpha \|\delta y_n^* - \delta y_n \|_2^2 + \mathcal{L}_C(l_n^*, l_n) \right ],
\label{eqn:m0}
\end{equation}
and its one-delta-per-pose-bin variant,
\begin{equation}
\M_S+: \min_W \frac{1}{N} \sum_n \left [ \alpha \mathcal{L}_C(l_n^*, l_n) + \|\delta y_n^* - \delta y_n^{\textcolor{red}{l_n}} \|_2^2 \right ].
\label{eqn:m0+}
\end{equation}
As discussed in \cite{Mahendran:ICCVW17}, the geodesic loss function is highly non-convex with many local optima, which makes it important to initialize the network weights correctly. We initialize weights of the Geodesic Bin \& Delta model and the Riemannian Bin \& Delta models by training the Simple Bin \& Delta model for one epoch of training data. These Simple Bin \& Delta models have been used in prior work \citep{Li:arxiv18,Guler:CVPR17} and they explicitly enforce supervision on the outputs of the two individual networks. 

\subsubsection{Log-Euclidean Bin \& Delta Model}
\label{sec:lebd}
The geodesic loss between the ground-truth and predicted rotations in the Riemannian Bin \& Delta model can be approximated by the Euclidean distance on the tangent space at the identity, 
\begin{align}
\mathcal{L}(R_n^*, \tilde{R}_{l_n} \exp(\delta y_n)) &= \mathcal{L}(\tilde{R}_{l_n}^T R_n^*, \exp(\delta y_n)) \nonumber \\ &\approx \| \log (\tilde{R}_{l_n}^T R_n^*)-\log (\exp (\delta y_n)) \|_2.
\end{align}
This approximation is better the closer $R_{l_n}$ is to $R_n^*$ or alternately, the closer $R_{l_n}^T R_n^*$ is to the Identity. This new regression loss gives us the Log-Euclidean Bin \& Delta model, 
\begin{equation}
\M_{LE}: \min_W \frac{1}{N} \sum_n \Big [ \mathcal{L}_c(l_n^*, l_n) + \alpha  \|\log(\tilde{R}_{l_n}^T R_n^*) - \delta y_n \|_2^2 \Big ] ,
\label{eqn:m2_old} 
\end{equation}
and its variant,
\begin{equation}
\M_{LE}+: \min_W \frac{1}{N} \sum_n \left [ \mathcal{L}_c(l_n^*, l_n) + \alpha \|\log(\tilde{R}_{l_n}^T R_n^*) - \delta y_n^{\textcolor{red}{l_n}} \|_2^2 \right ],
\label{eqn:m2+_old}
\end{equation}
where the term $\log(\tilde{R}_{l_n}^T R_n^*)$ can be precomputed for efficiency of training.

\section{Data Augmentation using 3D Pose Jittering}
\label{sec:data_augmentation}
\begin{figure}[h]
	\centering
	\begin{subfigure}{0.31\linewidth}
		\includegraphics[height=2cm, width=\linewidth]{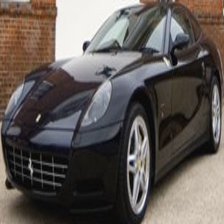}
		\caption{original}
	\end{subfigure}
	~
	\begin{subfigure}{0.31\linewidth}
		\includegraphics[height=2cm, width=\linewidth]{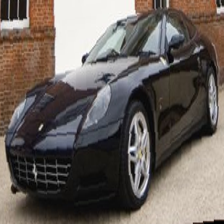}
		\caption{$\delta ct: +4^\circ$}
	\end{subfigure}
	~
	\begin{subfigure}{0.31\linewidth}
		\includegraphics[height=2cm, width=\linewidth]{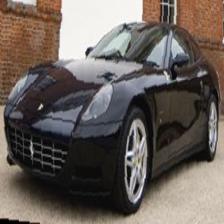}
		\caption{$\delta ct: -4^\circ$}
	\end{subfigure}
	
	\begin{subfigure}{0.31\linewidth}
		\includegraphics[height=2cm,width=\linewidth]{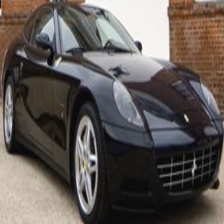}
		\caption{flipped}
	\end{subfigure}
	~
	\begin{subfigure}{0.31\linewidth}
		\includegraphics[height=2cm, width=\linewidth]{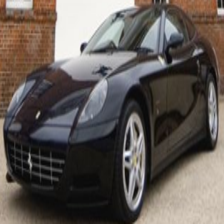}
		\caption{$\delta az: +2^\circ$}
	\end{subfigure}
	~
	\begin{subfigure}{0.31\linewidth}
		\includegraphics[height=2cm, width=\linewidth]{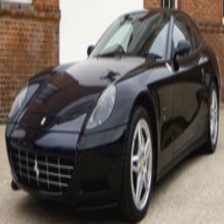}
		\caption{$\delta az: -2^\circ$}
	\end{subfigure}
	\caption{Augmented training samples from a car image}
	\label{fig:data_augmentation}
\end{figure}

We assume that each image is annotated with a 3D rotation $R(az, el, ct) = R_Z(ct) R_X(el) R_Z(az)$, where $R_Z$ and $R_X$ denote rotations around the $z$- and $x$-axis respectively. Jittered bounding boxes (bounding boxes with translational shifts that have sufficient overlap with the original box), like in V\&K \citep{Tulsiani:CVPR15}, introduce small unknown changes in the corresponding $R$. Instead, we augment our data by generating new samples corresponding to known small shifts in camera-tilt and azimuth. We call this new augmentation strategy \emph{3D pose jittering} (see Fig.~\ref{fig:data_augmentation}). Small shifts in camera-tilt lead to in-plane rotations, which are easily captured by rotating the image. Small shifts in azimuth or elevation angles lead to out-of-plane rotations, which can be described by homographies. We estimate these homographies in the following way: (i) we first project a CAD model of the object onto the image (we use a small percentage of the 3D points closest to the camera for the projection step), (ii) we rotate the object by a small angle in azimuth and/or elevation, (iii) we project the same set of points used earlier onto the image with new rotation (this is also the updated target for our training), and (iv) we compute a homography using the DLT algorithm \citep{Hartley-Zisserman04} between the two sets of projected points. We generate a dense grid of samples corresponding to $R(az \pm \delta az, el, ct \pm \delta ct)$. We also flip all samples, which corresponds to $R(-az, el, -ct)$. 

Along with these augmented images, we also use rendered images provided publicly by Render-for-CNN \citep{Su:ICCV15} \footnote{https://shapenet.cs.stanford.edu/media/\\syn\_images\_cropped\_bkg\_overlaid.tar}  to supplement our training data. 

\section{Results and Discussion}
\label{sec:results}
First, we describe the Pascal3D+ dataset \citep{Xiang:WACV14}, a popular benchmark dataset used for evaluating 3D pose estimation methods. Then we present the metrics we use to evaluate all our models. Then, we demonstrate the effectiveness of our framework and models with state-of-the-art performance on this challenging task. Finally, we present an ablation study on different decision choices we make in our pose estimation system.

\subsection{Dataset}
\label{sec:data}
The Pascal3D+ consists of images of twelve object categories: aeroplane (aero), bicycle (bike), boat, bottle, bus, car, chair, diningtable (dtable), motorbike (mbike), sofa, train and tvmonitor (tv). These images were curated from the Pascal VOC 2012 \cite{PASCAL} and ImageNet \cite{ImageNet} datasets, and annotated with 3D pose in terms of the Euler angles $(az, el, ct)$. We use the ImageNet-trainval as our training data,
Pascal-train images as our validation data and the Pascal-val images as our testing data. Following the protocol of \cite{Tulsiani:CVPR15,Su:ICCV15} and others, we use ground-truth bounding boxes of un-occluded and un-truncated objects. All our results in \S\ref{sec:3d_pose_results} are obtained using these ground-truth bounding boxes. We also evaluate the performance of our models on bounding boxes returned by object detection systems in 
\S\ref{sec:results_detected}.

\subsection{Evaluation Metrics}
\label{sec:evaluation_metrics}
To evaluate the performance of our models, we use two standard metrics proposed in \cite{Xiang:WACV14}: $MedErr$ and $Acc_\frac{\pi}{6}$. $MedErr$ is the median angle error (in degrees) between ground-truth and predicted rotation,
\begin{equation}
$MedErr$ \doteq \operatorname*{median}_{n=1}^N \angle (R_n^*, R_n),
\end{equation}
where $\angle (R_n^*, R_n) = \big |\cos^{-1} \big ( \frac{tr(R_n^T R_n^*) - 1}{2}\big ) \big |$ is the angle between ground-truth rotation $R_n^*$ and predicted rotation $R_n$ for test image $x_n$ \eqref{eqn:Rodrigues}.
$Acc_\frac{\pi}{6}$ is the percentage of test images that have angle error less than $30^\circ$
\begin{equation}
Acc_\frac{\pi}{6} \doteq \frac{1}{N} \sum_{n=1}^{N} \mathbf{1} \left[ \angle (R_n^*, R_n) < 30^\circ \right].
\end{equation}
To evaluate the performance of 3D pose estimation models with detected bounding boxes instead of ground-truth ones, \cite{Xiang:WACV14} and \cite{Tulsiani:CVPR15} extended the $AP$ (Average Precision) metric popularly used in the object detection literature to two metrics: $ARP_\theta$ and $AVP_K$. While computing the $AP$ metric, a detected bounding box is considered good if it has an intersection over union ($IOU$) overlap of at least 0.5 with a ground-truth box. For the $ARP$ metric, the detected bounding box must have sufficient overlap $IOU > 0.5$ \emph{and} have estimated rotation within $30^\circ$ of the ground-truth rotation, $ \angle (R_n^*, R_n) < 30^\circ$.
Note that we choose $\theta = 30^\circ$ similar to \cite{Tulsiani:CVPR15}. For the $AVP_K$ metric, the azimuth angles are binned into $K$ non-overlapping bins and a detection is considered good if $IOU > 0.5$ and $\ell(az^*) == \ell(az)$ where $\ell(az)$ is the bin corresponding to azimuth angle $az$. Previous works \citep{Tulsiani:CVPR15,Su:ICCV15,Massa:BMVC16} have reported performance using the $AVP_4$, $AVP_8$, $AVP_{16}$ and $AVP_{24}$ metrics. We also report our performance on these numbers. However, note that this metric is unfair for our models because whereas all previous works enforce supervision on the Euler angles (the azimuth angle included) directly, we enforce supervision on rotation matrices. We compute the azimuth angle from our estimated rotations and compare with ground-truth.

\subsection{3D Pose Estimation}
\label{sec:3d_pose_results}
We break down all our results along the lines of our models. We first discuss the performance of our pure regression models, then the classification model and finally our mixed-classification regression models. For all models discussed in this section and in \S\ref{sec:results_detected}, we use the axis-angle representation for rotation matrices. We discuss the choice of representation: axis-angle v/s quaternion as part of our ablation experiments in \S\ref{sec:ablation}. For all experiments and models we discuss in the next sections, we ran each experiment three times and report the mean across three trials.
For the error bars shown in all future figures, we use the standard deviation computed across these trials. 

\myparagraph{Pure Regression} Our first comparison is between the Geodesic and Euclidean regression models defined in \S\ref{sec:geodesic_regression} and \S\ref{sec:euclidean_regression} respectively. As can be seen in Table~\ref{table:regression}, using a geodesic regression loss significantly improves performance compared to Euclidean regression loss.
Table~\ref{table:regression_detailed} 
contains more detailed results with a breakdown per object category and the same behavior is observed for all object categories. 
This is in line with our expectations that a geodesic loss better reflects the underlying geometry of the problem and using this loss results in a big performance boost.

\begin{table}
	\centering
	\begin{tabular}{|c|cc|}
		\hline
		Model & $MedErr$ & $Acc_\frac{\pi}{6}$ \\
		\hline
		$\R_E$ & 15.50 & 0.7656 \\
		$\R_G$ & \textbf{11.63} & \textbf{0.8166} \\
		\hline
	\end{tabular}
\caption{Performance of our Euclidean and Geodesic Regression models on the test set. Lower is better for the $MedErr$ metric and higher is better for the $Acc_\frac{\pi}{6}$ metric.}
\label{table:regression}
\end{table}

\myparagraph{Pure Classification} We now evaluate the performance of the pure classification model described in \S\ref{sec:classification}. We ran K-Means clustering on the pose-targets of the rendered images to generate a pose dictionary. The size of the pose dictionary $K$ is a hyper-parameter of the model and we evaluated different models on the validation set to determine best choice of $K$. As can be seen in Table~\ref{table:hyperparameterK}, $K=200$ gave us the best results and is the size of the pose dictionary for all our mixed classification-regression models unless mentioned otherwise. In Table~\ref{table:classification}, we report the performance of our classification models on the test set and it is interesting to note that with a larger pose dictionary, we get a big improvement in the $MedErr$ metric for a reduction in $Acc_\frac{\pi}{6}$ performance.
We also see that the classification model seems to perform better than the regression model which is another motivation for our mixed classification-regression models.

\begin{table}
	\centering
	\begin{tabular}{|c|cc|}
		\hline
		$K$ & $MedErr$ & $Acc_\frac{\pi}{6}$ \\
		\hline
		50 & 15.68 & 0.8127 \\
		100 & 13.64 & 0.8170 \\
		200 & \textbf{12.50} & \textbf{0.8206} \\
		400 & 12.85 & 0.8162 \\
		\hline
	\end{tabular}
\caption{Performance of our Classification models on the validation set for different values of the size of the pose dictionary, $K$.}
\label{table:hyperparameterK}
\end{table}

\begin{table}
	\centering
	\begin{tabular}{|c|cc|}
		\hline
		Model & $MedErr$ & $Acc_\frac{\pi}{6}$ \\
		\hline
%		$\mathcal{C}$ ($K=100$) & 12.20 & \textbf{0.8350} \\
		$\mathcal{C}$ & \textbf{11.31} & \textbf{0.8298} \\
		\hline
		$\R_G$ & 11.63 & 0.8166 \\
		\hline
	\end{tabular}
\caption{Performance of our Classification model on the test set for $K = 200$ and a comparison with the Geodesic regression model.} 
\label{table:classification}
\end{table}

\begin{figure}
	\centering
	\includegraphics[width=\linewidth]{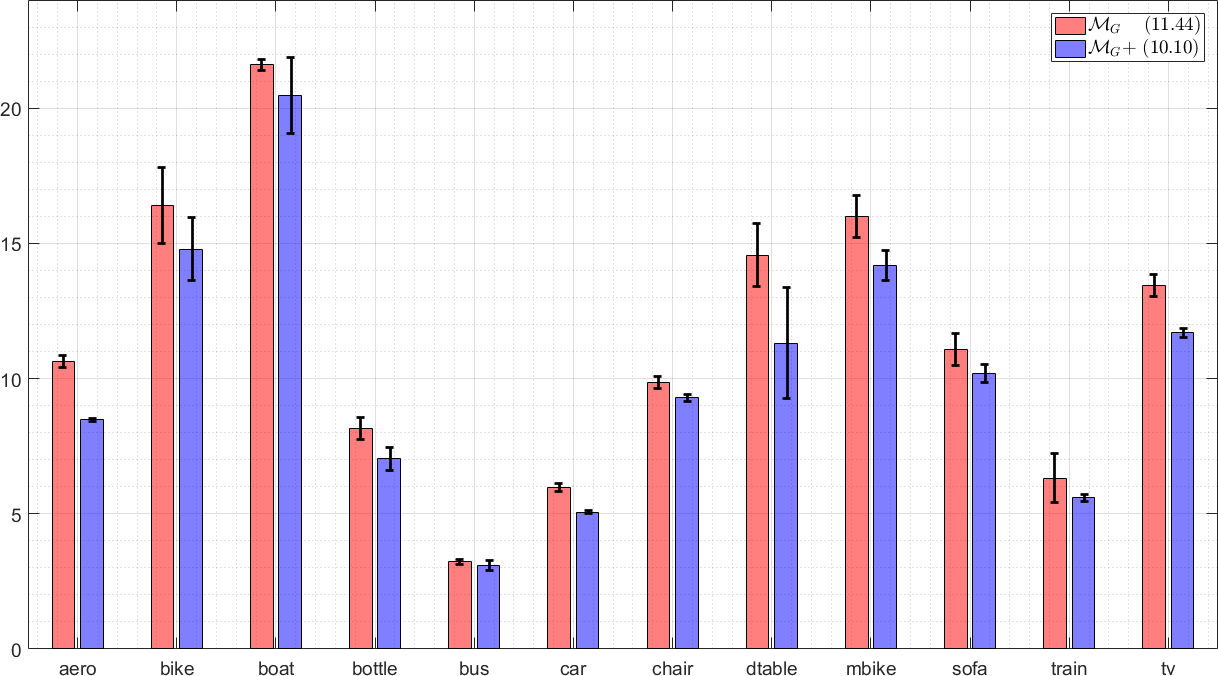}
	\caption{Performance of the Geodesic Bin \& Delta models per-category under the $MedErr$ metric}
	\label{fig:gbd_mederr}
\end{figure}

\myparagraph{Geodesic Bin \& Delta} We start with our first mixed classification-regression model, the Geodesic Bin \& Delta model (GBD in short), described in \S\ref{sec:gbd}. In our previous work \citep{Mahendran:arxiv18}, we noted that hyper-parameter values of $\alpha=1$ for $K=100$ for the $\mathcal{M}_G$ model and $\alpha=10$ for $K=16$ for the $\mathcal{M}_G+$ model worked best. For the $\mathcal{M}_G+$ model, we stay with this choice of hyper-parameters and study the choice of  $\alpha$ for the new choice of $K=200$ for the $\mathcal{M}_G$ model in Table~\ref{table:hyperparameteralpha}. We then report the performance of these models on the test set in Table~\ref{table:gbd_results} and Figure~\ref{fig:gbd_mederr}. As can be seen in the results, the one-delta-per-bin model in $\mathcal{M}_G+$ performs better for both metrics. Also, note that some object categories like diningtable, boat, bicycle and motorbike show larger errorbars compared to others and it is important to compare these and not just the absolute numbers to determine if the improvement a model makes is statistically significant. A comparison between the current-state-of-the-art pose estimation methods and a few of our models is shown in Table~\ref{table:state-of-the-art} and our Geodesic Bin \& Delta model $\mathcal{M}_G+$ achieves the state-of-the-art performance averaged across all twelve categories of the Pascal3D+ dataset under both the $MedErr$ and $Acc_\frac{\pi}{6}$ metrics. 

\begin{table}
	\centering
	\begin{tabular}{|c|cc|}
		\hline
		$\alpha$ & $MedErr$ & $Acc_\frac{\pi}{6}$ \\
		\hline
		0.1 & 12.83 & 0.8145 \\
		1.0 & \textbf{11.92} & \textbf{0.8212} \\
		10.0 & 13.90 & 0.8068 \\
		\hline
	\end{tabular}
\caption{Performance of our Geodesic Bin \& Delta model $\M_G$ on the validation test for different values of relative weighting parameter, $\alpha$, for $K=200$.}
\label{table:hyperparameteralpha}
\end{table}

\begin{table}
	\centering
	\begin{tabular}{|c|cc|}
		\hline
		Model & $MedErr$ & $Acc_\frac{\pi}{6}$ \\
		\hline
		$\M_G$ & 11.44 & 0.8439 \\
		$\M_G+$ & \textbf{10.10} & \textbf{0.8588} \\
		\hline
	\end{tabular}
\caption{Performance of the Geodesic Bin \& Delta models, $\M_G (\alpha=1, K=200)$ and $ \M_G+ (\alpha=10, K=16)$, on the test set.}
\label{table:gbd_results}
\end{table}

\begin{figure}
	\centering
	\includegraphics[width=\linewidth]{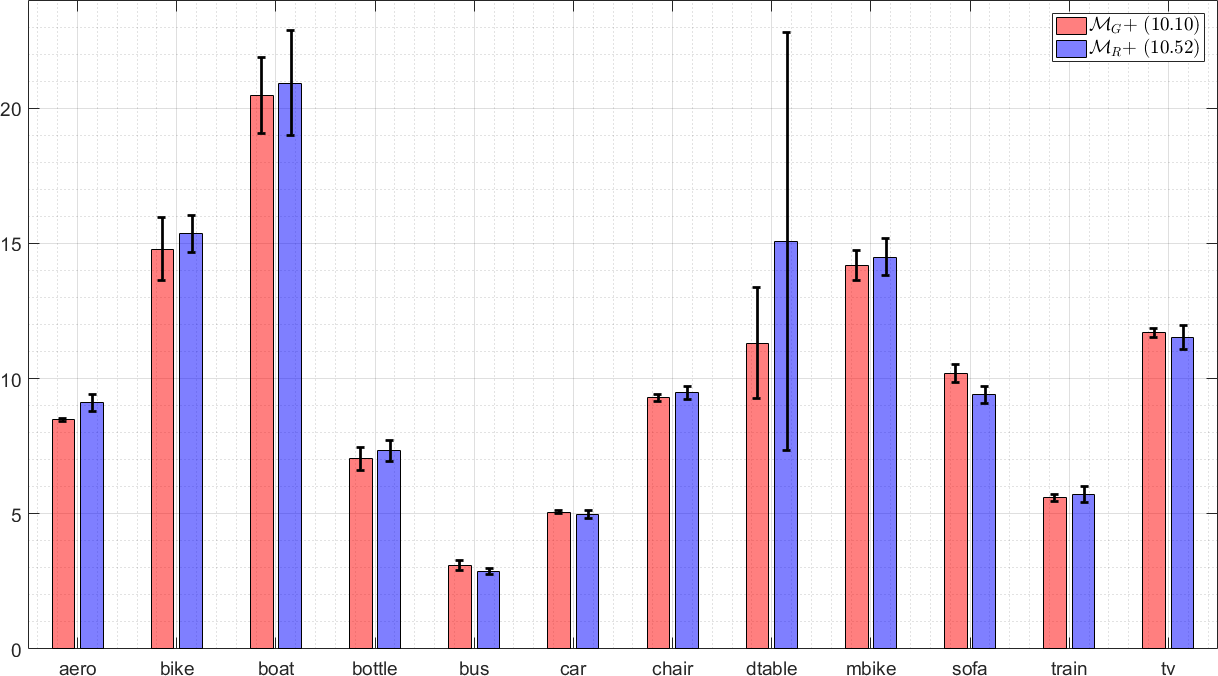}
	\caption{Comparison between GBD ($\M_G+$) and RBD ($\M_R+$) models under the $MedErr$ metric}
	\label{fig:rbd_mederr}
\end{figure}

\myparagraph{Riemannian Bin \& Delta} These models (RBD in short) were described in \S\ref{sec:rbd} and the key difference between these models and the GBD models discussed above is in what the pose residual $\delta y$ represents in both models. We use the same hyper-parameters of the GBD models for all future models including the Riemannian ones.
We report their performance in Table~\ref{table:rbd_results}. We also compare the performance of the RBD model $\mathcal{M}_R+$ model with the corresponding GBD model $\mathcal{M}_G+$ in Figure~\ref{fig:rbd_mederr}. A closer look shows that these two models are equivalent (with largely overlapping errorbars) for 8 out of the 12 object categories (bicycle, boat, bottle, car, chair, motorbike, train and tvmonitor). Also, note that the diningtable category where the $\M_R+$ model ($15.1^\circ$) is numerically much worse than the $\M_G+$ model ($11.3^\circ$) has a large errorbar due to bad performance for one of the three trials (diningtable $MedErr: 9.77^\circ, 9.43^\circ, 26.02^\circ$).

\begin{table}
	\centering
	\begin{tabular}{|c|cc|}
		\hline
		Model & $MedErr$ & $Acc_\frac{\pi}{6}$ \\
		\hline
		$\M_R$ & 11.69 & 0.8285 \\
		$\M_R+$ & \textbf{10.52} & \textbf{0.8573} \\
		\hline
	\end{tabular}
\caption{Performance of the Riemannian Bin \& Delta models $\M_R (\alpha=1, K=200)$ and $\M_R+ (\alpha=10, K=16)$, on the test set.}
\label{table:rbd_results}
\end{table}

\myparagraph{Probabilistic Bin \& Delta} These models (PBD in short) are described in \S\ref{sec:pbd} and are a probabilistic variation of the GBD models with the geodesic regression loss between ground-truth and predicted pose now weighted by the probability of pose-class predictions rather than just the most-likely pose-class. As can be seen in Table~\ref{table:pbd_results}, these models perform slightly worse than the GBD models but are still competitive. The advantage of these models is that these did not require any initialization strategy unlike GBD where we needed to initialize the models with 1 epoch of training the Simple Bin \& Delta models. 

\begin{table}
	\centering
	\begin{tabular}{|c|cc|}
		\hline
		Model & $MedErr$ & $Acc_\frac{\pi}{6}$ \\
		\hline
		$\M_P$ & 11.50 & \textbf{0.8491} \\
		$\M_P+$ & \textbf{10.80} & 0.8457 \\
		\hline
	\end{tabular}
\caption{Performance of the Probabilistic Bin \& Delta models, $\M_P(\alpha=1, K=200)$ and $\M_P+(\alpha=10, K=16)$, on the test set.}
\label{table:pbd_results}
\end{table}

\myparagraph{RelXed Bin \& Delta} These models (XBD in short) are also described in \S\ref{sec:xbd} and involve relaxing the hard-assignment of K-Means to a soft-assignment which also modifies the standard cross-entropy loss of the classification task to a Kullback-Liebler divergence between ground-truth and predicted probability distributions over the pose dictionary. A key hyper-parameter for these models is the $\gamma$ in Eqn.~\eqref{eqn:soft} which controls how peaky the probability distribution looks like. If we choose too high a $\gamma$ it will be equivalent to a hard-assignment, while if we choose too low a $\gamma$ it will lead to confusion between nearby pose classes. $\gamma$ is also a function of the pose dictionary and for the XBD models, we choose
\begin{equation}
\gamma = 0.5 \left [ \min_{i \neq j} \|z_i - z_j\|_2^2  \right ]^{-1}.
\end{equation}
This is can also be considered to be a function of the size of the pose dictionary, $K$. In Table~\ref{table:hyperparametergamma}, we see that increasing $K$ leads to worse results. We speculate that this is because as we increase $K$, we are discretizing the pose space into smaller clusters which are now closer to each other and more sensitive to both the choice of $\gamma$ and not as robust to mistakes in the classification task. 
For $K=16$, we just report performance for the one-delta-per-bin version of the model \ie for model $\mathcal{M}_X+: MedErr=11.53$ and $Acc_\frac{\pi}{6}=0.8407$. 

\begin{table}
	\centering
	\begin{tabular}{|c|c|cc|}
		\hline
		K & $\gamma$ & $MedErr$ & $Acc_\frac{\pi}{6}$ \\
		\hline
		16 & 2.06 & \textbf{12.48} & 0.8302 \\
		50 & 7.82 & 12.98 & \textbf{0.8320} \\
		100 & 15.08 & 51.71 & 0.3932 \\
		200 & 25.23 & 46.73 & 0.4236 \\
		\hline
	\end{tabular}
\caption{Performance of the RelaXed Bin \& Delta model $\M_X$ on the validation set for different choices of the size of pose dictionary, $K$.}
\label{table:hyperparametergamma}
\end{table}

\begin{figure}
	\centering
	\includegraphics[width=\linewidth]{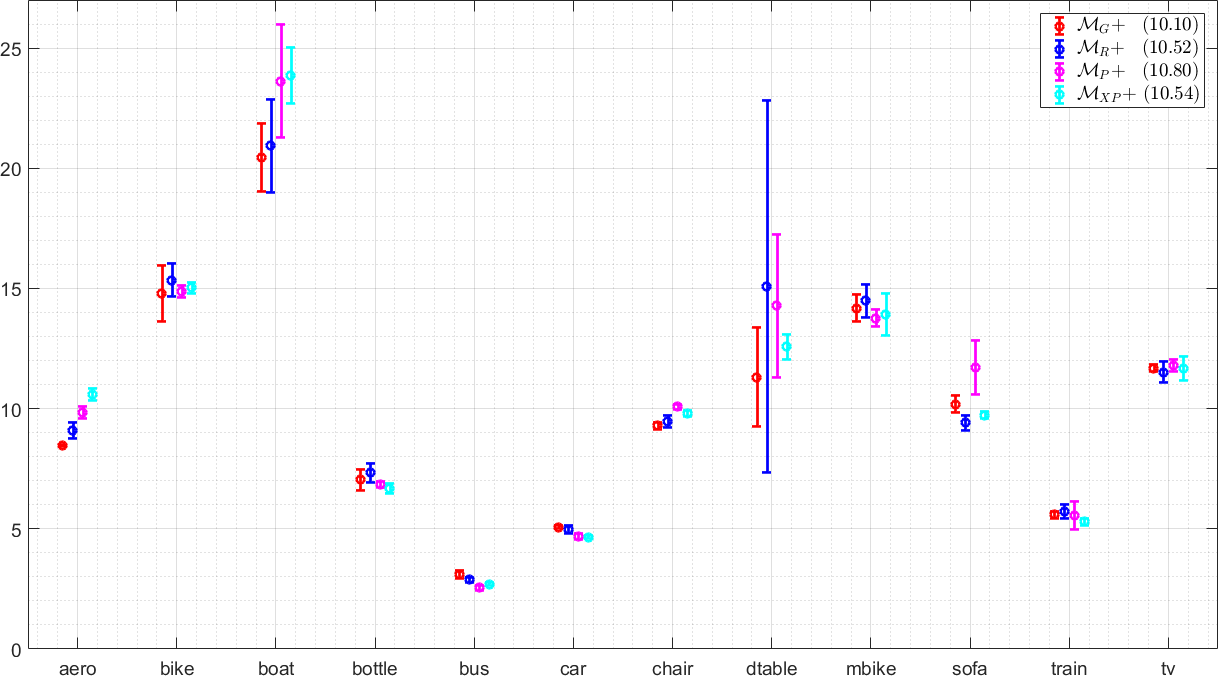}
	\caption{Comparison between GBD ($\M_G+$), RBD ($\M_R+$), PBD ($\M_P+$) and XPBD ($\M_{XP}+$) models under the $MedErr$ metric.}
	\label{fig:xpbd_mederr}
\end{figure}

\myparagraph{RelaXed Probabilistic Bin \& Delta} These models (XPBD in short) combine both the probabilistic weighting of geodesic losses and soft-assignment of pose-class labels. They were discussed in our previous work \citep{Mahendran:arxiv18}
as Probabilistic Bin \& Delta models and we report their performance in Table~\ref{table:xpbd_results}. Compared to the conference version of this work \citep{Mahendran:arxiv18},
where we combined both probabilistic relaxations into one model, a separate model for each modification shows us that the probabilistic weighting drives model performance and not the soft assignment. Fig.~\ref{fig:xpbd_mederr} shows a comparison of four of our best models ($\M_G+$, $\M_R+$, $\M_P+$ and $\M_{XP}+$) for all 12 object categories. We can see that for some categories like bicycle, bottle, bus, diningtable, motorbike, train and tvmonitor, the errorbars for all four models overlap and can be considered equivalent. For some categories like bus, boat, car and chair, two pairs of equivalent models, $\M_G+$ \& $\M_R+$ and $\M_P+$ \& $\M_{XP}+$ are formed where one pair is better than the other. 

\begin{table}
	\centering
	\begin{tabular}{|c|cc|}
		\hline
		Model & $MedErr$ & $Acc_\frac{\pi}{6}$ \\
		\hline
		$\M_{XP}$ & 11.38 & 0.8185 \\
		$\M_{XP}+$ & \textbf{10.54} & \textbf{0.8470} \\
		\hline
	\end{tabular}
\caption{Performance of the RelaXed Probabilistic Bin \& Delta models, $\M_{XP}(\alpha=1, K=100, \gamma=10)$ and $\M_{XP}+(\alpha=1, K=16, \gamma=10)$ on the test set.}
\label{table:xpbd_results}
\end{table}

\myparagraph{Simple Bin \& Delta and Log-Euclidean Bin \& Delta} We report the performance of these models in Table~\ref{table:sbd_lebd_results}. Based on our previous work \citep{Mahendran:ICCVW17}, we observed that applying a Euclidean regression loss on just the output of the Delta network is not sufficient and is not geometrically accurate. We make an assumption that with a discretized pose space, the Euclidean or Log-Euclidean regression loss starts to approximate the Geodesic regression loss. However, this is either not observed in practice (as can be seen from our results) or is practically infeasible (we would need a one-delta-per-bin model with a much finer discretization). We did not study these models in more detail in this work. 

\begin{table}
	\centering
	\begin{tabular}{|c|cc|}
		\hline
		Model & $MedErr$ & $Acc_\frac{\pi}{6}$ \\
		\hline
		$\M_S$ & 12.14 & 0.8303 \\
		$\M_S+$ & 11.95 & 0.8387 \\
		$\M_{LE}$ & 12.11 & 0.8410 \\
		$\M_{LE}+$ & 11.99 & 0.8329 \\
		\hline
	\end{tabular}
	\caption{Performance of our Simple Bin \& Delta models, $\M_S (\alpha=1, K=100)$ and $\M_S+ (\alpha=1, K=16)$, and Log-Euclidean Bin \& Delta models, $\M_{LE} (\alpha=1, K=100)$ and $\M_{LE}+ (\alpha=1, K=16)$ on the test set.}
	\label{table:sbd_lebd_results}
\end{table}

\begin{figure*}
	\centering
	\begin{subfigure}{0.48\linewidth}
		\includegraphics[width=\linewidth]{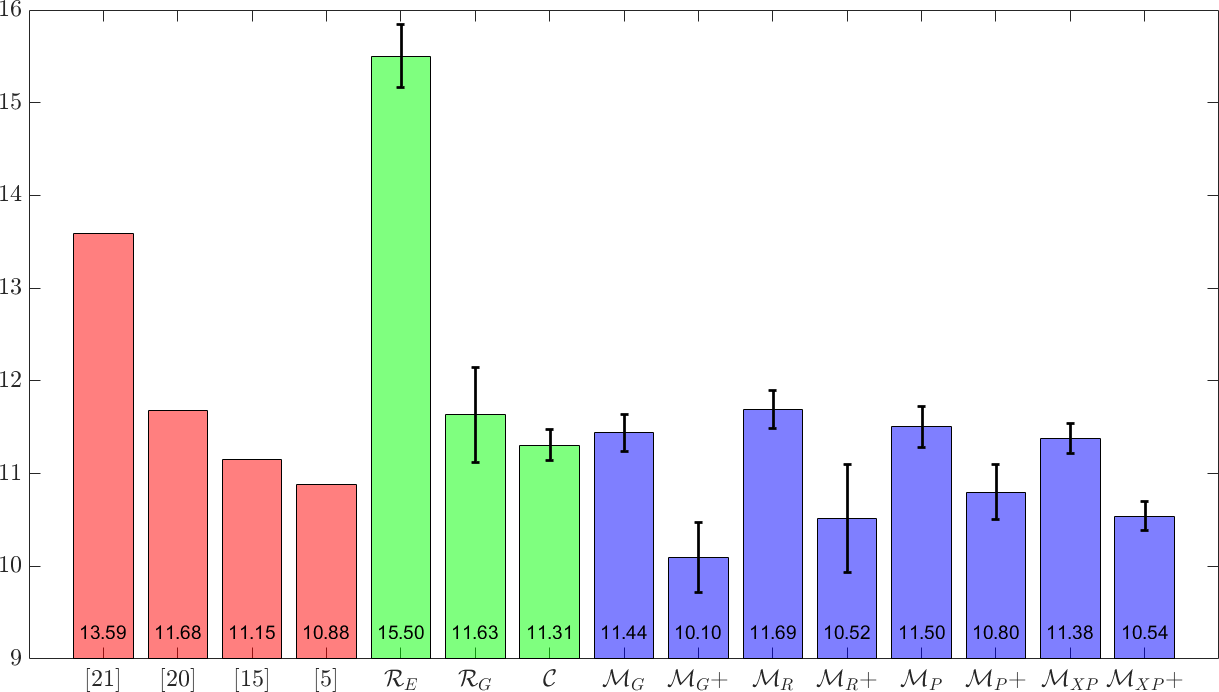}
		\caption{$MedErr$ metric}
	\end{subfigure}
	\hfill
	\begin{subfigure}{0.48\linewidth}
		\includegraphics[width=\linewidth]{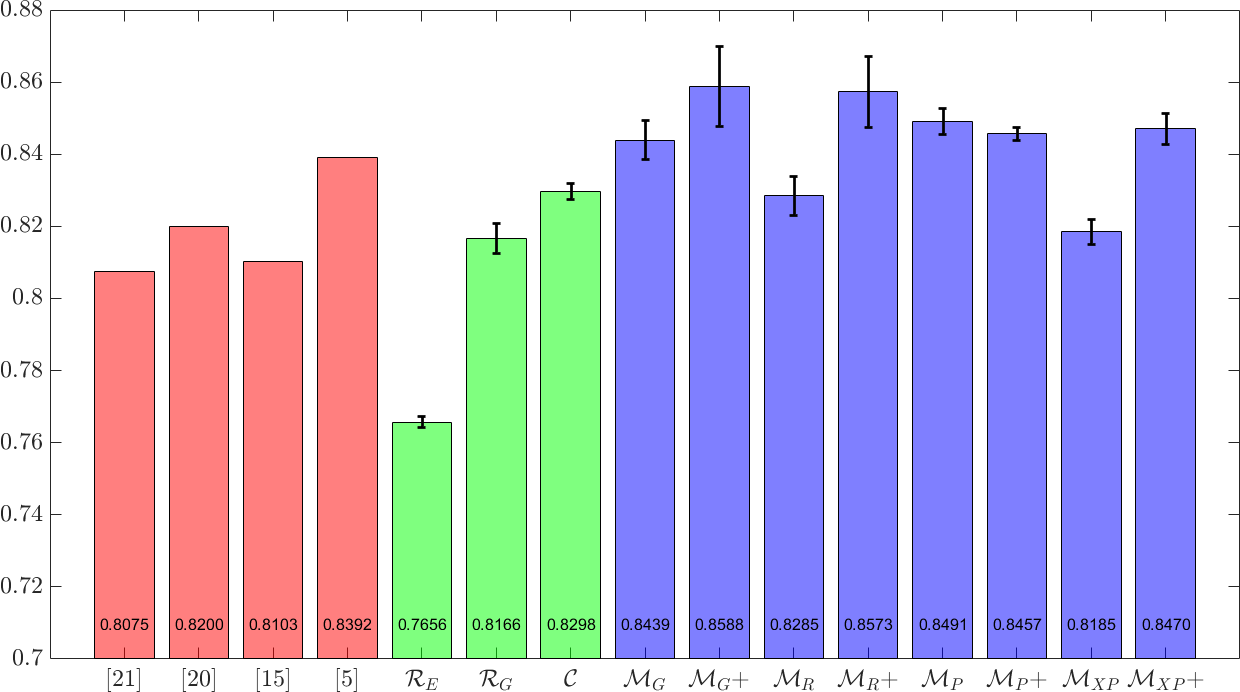}
		\caption{$Acc_\frac{\pi}{6}$ metric}
	\end{subfigure}
	\caption{Comparison of our models with state-of-the-art pose estimation methods under different metrics. Current state-of-the-art models are shown in red. The pure regression and classification models are shown in green. Our Bin \& Delta models are shown in blue. Lower is better.}
	\label{fig:sota_results}
\end{figure*}

\myparagraph{State-of-the-Art comparison} We compare our results with those of state-of-the-art methods in Table~\ref{table:state-of-the-art} and show that for four of our Bin \& Delta models achieve better performance and that our Geodesic Bin \& Delta model achieves the best results. In Fig.~\ref{fig:sota_results}, we compare the performance of our best mixed classification-regression models with some baseline pure regression, pure classification methods and the current state-of-the-art pose estimation methods averaged across all object categories and show that we achieve better results under both $MedErr$ and $Acc_\frac{\pi}{6}$ metrics. A full list of all our Bin \& Delta models with per-category performance under both metrics can be found in Table~\ref{table:all_bd_detailed}. 

\begin{table*}
	\centering
	\setlength{\tabcolsep}{1mm}
	\begin{tabular}{|c|c|cccccccccccc|c|}
		\hline
		Metric & Model & aero & bike & boat & bottle & bus & car & chair & dtable & mbike & sofa & train & tv & Mean \\
		\hline
		\multirow{8}{*}{$MedErr$} & \citep{Tulsiani:CVPR15} & 13.8 & 17.7 & 21.3 & 12.9 & 5.8 & 9.1 & 14.8 & 15.2 & 14.7 & 13.7 & 8.7 & 15.4 & 13.59 \\
		& \citep{Su:ICCV15} & 15.4 & \textcolor{red}{14.8} & 25.6 & 9.3 & 3.6 & 6.0 & 9.7 & \textbf{10.8} & 16.7 & \textcolor{red}{9.5} & 6.1 & 12.6 & 11.68 \\
		& \citep{Mousavian:CVPR17} & 13.6 & \textbf{12.5} & 22.8 & 8.3 & 3.1 & 5.8 & 11.9 & 12.5 & \textcolor{red}{12.3} & 12.8 & 6.3 & 11.9 & 11.15 \\
		& \citep{Grabner:CVPR18} & 10.0 & 15.6 & \textbf{19.1} & 8.6 & 3.3 & 5.1 & 13.7 & 11.8 & \textbf{12.2} & 13.5 & 6.7 & \textbf{11.0} & 10.88 \\
		\cline{2-15}
		& $\M_G+$ & \textbf{8.5} & \textcolor{red}{14.8} & \textcolor{red}{20.5} & 7.0 & 3.1 & 5.1 & \textbf{9.3} & \textcolor{red}{11.3} & 14.2 & 10.2 & \textcolor{red}{5.6} & 11.7 & \textbf{10.10} \\ 
		& $\M_R+$ & \textcolor{red}{9.1} & 15.3 & 20.9 & 7.3 & 2.9 & \textcolor{red}{5.0} & \textcolor{red}{9.5} & 15.1 & 14.5 & \textbf{9.4} & 5.7 & \textcolor{red}{11.5} & \textcolor{red}{10.52} \\ 
		& $\M_P+$ & 9.8 & 14.9 & 23.6 & \textcolor{red}{6.8} & \textbf{2.5} & \textbf{4.7} & 10.1 & 14.3 & 13.8 & 11.7 & \textcolor{red}{5.6} & 11.8 & 10.80 \\ 
		& $\M_{XP}+$ & 10.6 & 15.0 & 23.9 & \textbf{6.7} & \textcolor{red}{2.7} & \textbf{4.7} & 9.8 & 12.6 & 13.9 & 9.7 & \textbf{5.3} & 11.7 & 10.54 \\ 		
		\hline
		\multirow{11}{*}{$Acc_\frac{\pi}{6}$} & \citep{Tulsiani:CVPR15} & 0.81 & 0.77 & 0.59 & 0.93 & \textbf{0.98} & 0.89 & 0.80 & 0.62 & \textbf{0.88} & 0.82 & 0.80 & 0.80 & 0.8075 \\
		& \citep{Su:ICCV15} & 0.74 & \textbf{0.83} & 0.52 & 0.91 & 0.91 & 0.88 & 0.86 & \textbf{0.73} & 0.78 & 0.90 & \textbf{0.86} & \textbf{0.92} & 0.8200 \\
		& \citep{Mousavian:CVPR17} & 0.78 & \textbf{0.83} & 0.57 & 0.93 & 0.94 & 0.90 & 0.80 & 0.68 & \textcolor{red}{0.86} & 0.82 & 0.82 & 0.85 & 0.8103 \\
		& \citep{Grabner:CVPR18} & 0.83 & \textcolor{red}{0.82} & \textbf{0.64} & 0.95 & \textcolor{red}{0.97} & \textcolor{red}{0.94} & 0.80 & \textcolor{red}{0.71} & \textbf{0.88} & 0.87 & 0.80 & 0.86 & 0.8392 \\
		\cline{2-15}
		& $\M_G+$ & \textbf{0.87} & 0.81 & \textbf{0.64} & \textcolor{red}{0.96} & \textcolor{red}{0.97} & \textbf{0.95} & \textbf{0.92} & 0.67 & 0.85 & \textbf{0.97} & 0.82 & 0.88 & \textbf{0.8588} \\ 
		& $\M_R+$ & \textcolor{red}{0.86} & 0.81 & \textcolor{red}{0.62} & \textcolor{red}{0.96} & \textcolor{red}{0.97} & \textbf{0.95} & \textbf{0.92} & 0.67 & 0.83 & \textbf{0.97} & \textcolor{red}{0.83} & \textcolor{red}{0.90} & \textcolor{red}{0.8573} \\ 
		& $\M_P+$ & 0.85 & 0.79 & 0.60 & \textcolor{red}{0.96} & \textcolor{red}{0.97} & \textbf{0.95} & \textcolor{red}{0.88} & 0.68 & 0.82 & \textcolor{red}{0.93} & 0.81 & 0.89 & 0.8457 \\ 
		& $\M_{XP}+$ & 0.84 & \textcolor{red}{0.82} & 0.59 & \textbf{0.97} & \textcolor{red}{0.97} & \textbf{0.95} & \textcolor{red}{0.88} & 0.68 & 0.84 & \textcolor{red}{0.93} & 0.81 & 0.89 & 0.8470 \\ 		
		\hline
	\end{tabular}
	\caption{Comparison with current state-of-the-art algorithms for 3D pose estimation from 2D images on the Pascal3D+ dataset under different metrics. Lower is better for the MedErr metric and higher is better for the Accuracy metric. Best results are highlighted in bold and second-best results are shown in red (best seen in color).}
	\label{table:state-of-the-art}
\end{table*}

\subsection{Ablation Analysis}
\label{sec:ablation}
In this section, we discuss three decision choices we made in our pose estimation pipeline: (1) Choice of representation, (2) Choice of feature network and (3) Choice of data augmentation. 

\begin{figure}
	\centering
	\includegraphics[width=\linewidth]{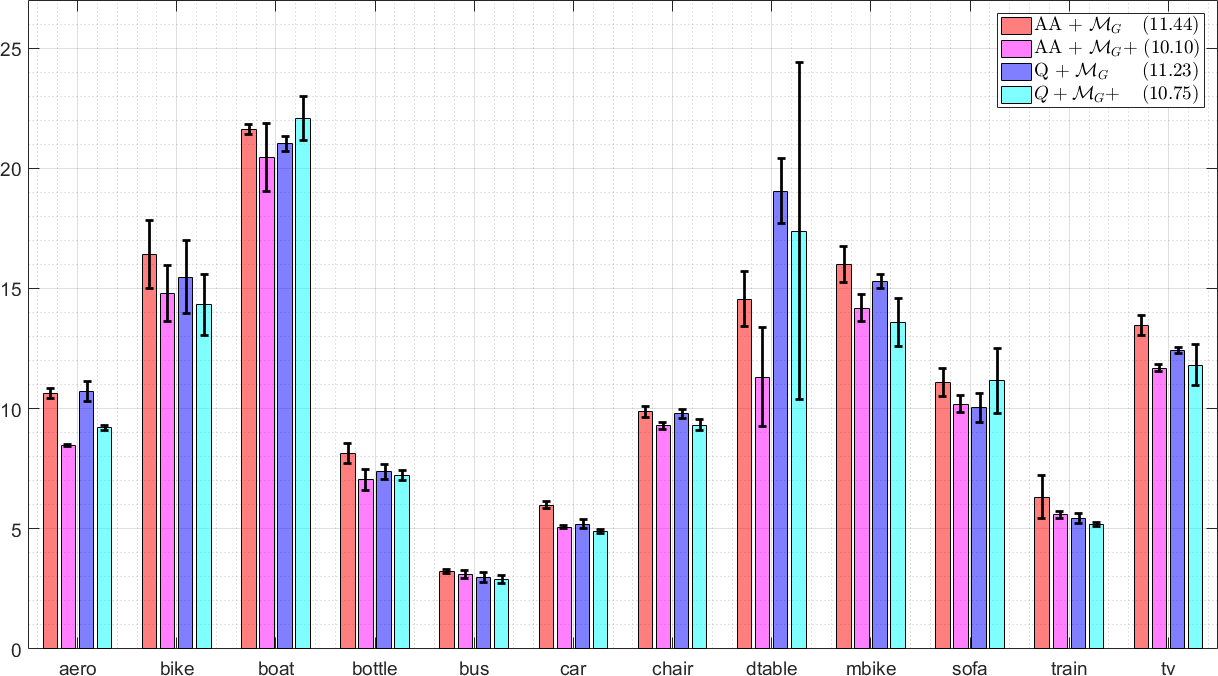}
	\caption{Comparison between Axis-angle and Quaternion representations under the $MedErr$ metric.}
	\label{fig:quaternion_mederr}
\end{figure}

\myparagraph{Orientation Representation} As we discussed earlier in \S\ref{sec:representations}, we can represent a rotation matrix in terms of Euler angles, Axis-angles or Quaternions. We prefer to use the axis-angle representation due to its compactness and the geodesic properties defined on top of it. However, quaternions could also have been chosen and all the models re-defined with them instead. We now study how the choice of quaternions as our pose representation would affect the performance of our models. We use quaternions in our Euclidean regression model $\R_E$, our Geodesic Regression model $\R_G$ and our Geodesic Bin \& Delta models $\R_G$ and $\R_G+$. As can be seen in Table~\ref{table:quaternion_results} and Fig.~\ref{fig:quaternion_mederr}, both representations are equivalent for most object categories with axis-angle being better for some. For some models like the Riemannian Bin \& Delta and the Log-Euclidean Bin \& Delta, the axis-angle representation is a more direct choice compared to quaternions. 

\begin{table}
	\centering
	\begin{tabular}{|cc|cc|}
		\hline
		Model & Representation & $MedErr$ & $Acc_\frac{\pi}{6}$ \\
		\hline
		\multirow{2}{*}{$\R_E$} & Axis-angle & 15.50 & 0.7656 \\
		& Quaternion & 14.14 & 0.7965 \\
		\hline
		\multirow{2}{*}{$\R_G$} & Axis-angle & 11.63 & 0.8166 \\
		& Quaternion & 12.20 & 0.8141 \\
		\hline
		\multirow{2}{*}{$\M_G$} & Axis-angle & 11.44 & 0.8439 \\
		& Quaternion & 11.23 & 0.8384 \\
		\hline
		\multirow{2}{*}{$\M_G+$} & Axis-angle & \textbf{10.10} & \textbf{0.8588} \\
		& Quaternion & 10.75 & 0.8560 \\
		\hline
	\end{tabular}
\caption{Performance of Axis-angle and Quaternion pose representations on the test set under different models.}
\label{table:quaternion_results}
\end{table}

\myparagraph{Feature Network} In our first work on 3D pose regression \citep{Mahendran:ICCVW17}, we used the VGG-M network \citep{Chatfield:BMVC14} and in our recent work, we use the ResNet50 network \citep{He:CVPR16}. One could exhaustively search over all possible feature networks and find the one that performs the best but that is not the aim of this work. Instead, we just analyze two choices of feature networks: a standard VGG13 with Batch-Norms (from the PyTorch model zoo) and a ResNet50 (also from the PyTorch model zoo) and compare their performance under our Geodesic Bin \& Delta models. As can be seen in Table~\ref{table:feature_network}, the models with a ResNet feature network perform better than the models with a VGG feature network. This has also been observed previously in other works  \citep{Grabner:CVPR18}.

\begin{table}
	\centering
	\begin{tabular}{|cc|cc|}
		\hline
		Model & Network & $MedErr$ & $Acc_\frac{\pi}{6}$ \\
		\hline
		\multirow{2}{*}{$\M_G$} & ResNet50 & 11.44 & 0.8439 \\
		& VGG13 & 11.83 & 0.8277 \\
		\hline
		\multirow{2}{*}{$\M_G+$} & ResNet50 & \textbf{10.10} & \textbf{0.8588} \\
		& VGG13 & 10.81 & 0.8423 \\
		\hline
	\end{tabular}
\caption{Comparison between ResNet50 and VGG13 feature networks with the Geodesic Bin \& Delta models.}
\label{table:feature_network}
\end{table}

\myparagraph{Data Augmentation} We use both 3D Pose jittered and rendered images to augment our training data. We now study how using just one of these would affect the performance of our Geodesic Bin \& Delta models. As can be seen in Table~\ref{table:data_augmentation}, using both augmented and rendered images is important to train our models. We believe that the more powerful model $\mathcal{M}_G+$ performs worse than $\mathcal{M}_G$ when training on purely rendered images because it is overfitting rendered data which is different from the real data distribution/images on which the models are finally tested. 

\begin{table}
	\centering
	\begin{tabular}{|cc|cc|}
		\hline
		Model & Data & $MedErr$ & $Acc_\frac{\pi}{6}$ \\
		\hline
		\multirow{3}{*}{$\M_G$} & Rendered & 14.79 & 0.7779 \\
		& Augmented & 18.35 & 0.6757 \\
		& Both & 11.44 & 0.8439 \\
		\hline
		\multirow{3}{*}{$\M_G+$} & Rendered & 15.03 & 0.7748 \\
		& Augmented & 14.30 & 0.7506 \\
		& Both & \textbf{10.10} & \textbf{0.8588} \\
		\hline
	\end{tabular}
\caption{Comparison between different types of data augmentation with the Geodesic Bin \& Delta models.}
\label{table:data_augmentation}
\end{table}

\subsection{3D Pose Estimation with Detected bounding boxes}
\label{sec:results_detected}
So far all the experimental evaluation was done with ground-truth bounding boxes with un-occluded and un-truncated objects. This allows us to focus on the orientation estimation problem assuming known detection and categorization. In this section, instead of ground-truth bounding boxes returned by an oracle, we use bounding boxes returned by an object detection system. As mentioned earlier in \S\ref{sec:evaluation_metrics}, we compare with other methods under the modified metrics for joint object detection and pose estimation: the $ARP_\frac{\pi}{6}$ and $AVP_K$ metrics where $K=4,8,16,24$. We again emphasize that the AVP metric is an unfair metric for our models because we do not enforce any supervision on the azimuth angle unlike other models. We still report these numbers for the sake of completeness. 

Firstly, as can be seen in Table~\ref{table:arp}, we significantly improve upon the state-of-the-art \citep{Tulsiani:CVPR15} under the $ARP_\frac{\pi}{6}$ metric. We evaluate the performance of our Geodesic Bin \& Delta models ($\M_G$ and $\M_G+$) with bounding boxes returned by three object detection systems: (i) RCNN\footnote{\url{http://www.cs.berkeley.edu/~shubhtuls/cachedir/vpsKps/VOC2012_val_det.mat}} \citep{Tulsiani:CVPR15} (ii) RCNN\footnote{\url{https://github.com/ShapeNet/RenderForCNN/tree/master/data/detection_results/rcnn_bbox_reg_pruned}} \citep{Su:ICCV15} and (iii) Mask-RCNN\footnote{X-101-64x4d-FPN backbone (model \#36494496): \url{https://github.com/facebookresearch/Detectron/blob/master/MODEL_ZOO.md}} \citep{He:ICCV17,Detectron2018}. A direct comparison between our work and that of \cite{Tulsiani:CVPR15} using their detected bounding boxes shows that we improve upon their performance by at least 5 points. This shows that we are actually improving the pose estimation performance and not just boosting the metric artificially using better object detection systems. Using the current state-of-the-art detection system of Mask RCNN leads to an even bigger improvement of around 15.1 points for the $\M_G$ model and 17.4 points for the $\M_G+$ model.
Additionally, we observe that the $\M_G+$ model is consistently better than the $\M_G$ model for all detection systems. This is consistent with the performance on ground-truth boxes where the  $\M_G+$ model ($10.10^\circ$) has lower median angle error compared to the $\M_G$ model ($11.44^\circ$). Similarly we observe that the ARP performance follows the order RCNN $<$ V\&K $<$ Mask-RCNN. This is also consistent with the object detection performance under the AP metric.
We analyze the performance of these models in more detail in Table~\ref{table:detection_analysis} and it shows that better detections is just part of the reason for improved performance. We report three metrics: (i) \% Detected: the percentage of bounding boxes detected correctly by the detection system (IoU overlap $>0.5$), (ii) \% Correct: the percentage of bounding boxes detected correctly and having pose error $<30^\circ$, and (iii) MedErr: median pose error (in degrees) for all detected bounding boxes. Note that Pose-Err includes even those cases where the bounding box was detected correctly but the pose was estimated incorrectly. From this table, we see that both \% Detected and \% Correct increase with a better detection system but also the Pose-Err goes down. This tells us that we are not only detecting more bounding boxes with Mask-RCNN + $\M_G+$ but also estimating their pose more accurately. 

\begin{table}
	\centering
	\begin{tabular}{|c|c|c|c|}
		\hline
		Model & Detections & $ARP_\frac{\pi}{6}$ & AP \\
		\hline
		V\&K & \multirow{3}{*}{V\&K} & 46.5 & \multirow{3}{*}{61.6}\\
		$\M_G$ & & 51.6 & \\
		$\M_G+$ & & \textcolor{red}{52.6} & \\
		\hline
		$\M_G$ & \multirow{2}{*}{R4CNN} & 50.2 & \multirow{2}{*}{61.0}\\
		$\M_G+$ & & 51.3 & \\
		\hline
		$\M_G$ & \multirow{2}{*}{Mask-RCNN} & 61.6 & \multirow{2}{*}{\textbf{76.2}}\\
		$\M_G+$ & & \textbf{63.9} & \\
		\hline
	\end{tabular}
\caption{Performance of our Geodesic Bin \& Delta models on the test using under the $ARP_\frac{\pi}{6}$ metric with different detected bounding boxes. Higher is better.}
\label{table:arp}
\end{table}

\begin{table}
	\setlength{\tabcolsep}{1mm}
	\centering
	\begin{tabular}{|cc|ccc|}
		\hline
		Model & Detections & \%Detected & \%Correct & Pose-Err \\
		\hline
		\multirow{3}{*}{$\M_G$} & V\&K & 0.7392 & 0.5347 & 16.06 \\
		& R4CNN & 0.7258 & 0.5264 & 16.01 \\
		& Mask-RCNN & \textbf{0.8158} & 0.6028 & 15.28 \\
		\hline
		\multirow{3}{*}{$\M_G+$} & V\&K & 0.7392 & 0.5469 & 14.96 \\
		& R4CNN & 0.7258 & 0.5406 & 14.77 \\
		& Mask-RCNN & \textbf{0.8158} & \textbf{0.6225} & \textbf{13.83} \\
		\hline
	\end{tabular}
\caption{Performance of the Geodesic Bin \& Delta models with detected bounding boxes. We report three metrics for each combination of model and detection system. Please see text for more details. Higher is better for the percentages and lower is better for Pose-Err.}
\label{table:detection_analysis}
\end{table}

As can be seen in Table~\ref{table:avp}, for the AVP metric, we are clearly better than \cite{Su:ICCV15} but are worse than \cite{Tulsiani:CVPR15} and \cite{Massa:BMVC16}, especially for a higher number of azimuth bins $K=24$. Note that \cite{Massa:BMVC16} solves a joint object detection and pose estimation problem within a Fast-RCNN \citep{Girshick:ICCV15} detection framework, which in principle should give better results than systems trained on ground-truth bounding boxes and evaluated on detected ones (like V\&K, R4CNN and Ours) 

\begin{table}
	\centering
	\setlength{\tabcolsep}{1mm}
	\begin{tabular}{|c|c|cccc|}
		\hline
		Model & Detections & $AVP_4$ & $AVP_8$ & $AVP_{16}$ & $AVP_{24}$ \\
		\hline
		V\&K & V\&K & 49.1 & 44.5 & 36.0 & 31.1 \\
		R4CNN & R4CNN & 39.7 & 32.9 & 24.2 & 19.8 \\
		Massa & Fast-RCNN & 55.4 & 51.3 & \textbf{40.6} & \textbf{36.1} \\
		\hline
		\multirow{3}{*}{$\M_G$} & V\&K & 49.0 & 41.5 & 31.5 & 25.2 \\
		& R4CNN & 48.2 & 40.2 & 29.2 & 24.4 \\
		& Mask-RCNN & 59.2 & 50.6 & 36.4 & 30.0 \\
		\hline
		\multirow{3}{*}{$\M_G+$} & V\&K & 49.9 & 43.1 & 33.1 & 27.3 \\
		& R4CNN & 49.4 & 42.5 & 32.0 & 26.3 \\
		& Mask-RCNN & \textbf{62.0} & \textbf{53.2} & 40.2 & 33.5 \\
		\hline
	\end{tabular}
\caption{Performance of our Geodesic Bin \& Delta models on the test using under the $AVP_K$ metric with different detected bounding boxes. Higher is better.}
\label{table:avp}
\end{table}

\section{Conclusion}
\label{sec:conclusion}
Orientation estimation is a challenging computer vision problem with applications in many domains. Current deep learning methods, in spite of their impressive performance, do not fully exploit the geometry of the orientation space. We designed regression-based deep learning models that use representations and loss functions that respect the Riemannian structure of the orientation space. We also built a framework of mixed classification-regression that leads to a family of Bin \& Delta models that model the geometry of the pose data. We trained these models using a data augmentation strategy that is designed to capture perturbations in the orientation space. Our proposed models achieved state-of-the-art results on the challenging PASCAL3D+ benchmarking dataset across a variety of metrics using both ground-truth and detected bounding boxes demonstrating the significant gain in performance we achieved by using geometrically appropriate models. 

\myparagraph{Acknowledgements} This research work was supported by NFS grants 1527340 and 1834427.
\appendix

\section{Implementation Details}
\label{sec:implementation_details}

All our code was implemented in PyTorch \cite{} and will be made publicly available \footnote{\url{https://github.com/JHUVisionLab/multi-modal-regression}} . We used Adam optimizer with an initial learning rate of $10^{-4}$ and subsequent reductions by a factor of $0.1$ after every epoch. For the One-bin-and-delta models: $\M_G$, $\M_R$, $\M_P$ and $\M_{XP}$, for every object category, the Bin network is of size 2048-1000-500-200 for choice of $K=100$ and the Delta network is of size 2048-1000-500-3. For the One-delta-per-bin models: $\M_G+$, $\M_R+$, $\M_P+$ and $\M_{XP}+$, for every object category, the Bin network is of size 2048-1000-500-16 for choice of $K=16$ and the Delta network is of size of 16 x 2048-100-3 as we have 16 delta networks in total (1 per bin). Different object categories have different number of images in the Pascal3D+ dataset and we use a sampling strategy to balance the data in every mini-batch. We sample 4 rendered images and 4 real (augmented) images per object category in every mini-batch leading to a batch size of 96 images. This was the largest batch size that fit into a single GPU (Titan Xp) and was chosen for all One-bin-and-delta models. For the One-delta-per-bin models, we used 3 rendered images and 3 real images per object category instead as the largest batch size that fit into a single GPU. We generated augmented data with azimuth shifts $[-1, 0, 1]$, elevation shifts $[-1, 0, 1]$ and camera-tilt shifts $[-4, -2, 0, 2, 4]$. 

\section{More Results}
\label{sec:more_results}
In \S\ref{sec:3d_pose_results} and \S\ref{sec:ablation}, we provided results for different models and experiments averaged across all twelve object categories. We now provide expanded tables with per-category results. We also provide a list of correspondences in Table~\ref{table:list_of_tables}.
\begin{table}[h]
	\centering
	\begin{tabular}{|cc|}
		\hline
		Condensed Results & Expanded Results \\
		\hline
		Table~\ref{table:regression} & Table~\ref{table:regression_detailed} \\
		Table~\ref{table:hyperparameterK} & Table~\ref{table:hyperparameterK_detailed} \\
		Table~\ref{table:classification} & Table~\ref{table:classification_detailed} \\
		Table~\ref{table:hyperparameteralpha} & Table~\ref{table:hyperparameteralpha_detailed} \\
		Tables~\ref{table:gbd_results},\ref{table:rbd_results},\ref{table:pbd_results},\ref{table:xpbd_results},\ref{table:sbd_lebd_results} & Table~\ref{table:all_bd_detailed} \\
		Table~\ref{table:hyperparametergamma} & Table~\ref{table:hyperparametergamma_detailed} \\
		Table~\ref{table:quaternion_results} & Table~\ref{table:representation_detailed} \\
		Table~\ref{table:data_augmentation} & Table~\ref{table:data_augmentation_detailed} \\
		Table~\ref{table:feature_network} & Table~\ref{table:feature_network_detailed} \\
		Table~\ref{table:arp} & Tables~\ref{table:arp_detailed},\ref{table:ap_detailed} \\
		Table~\ref{table:detection_analysis} & Table~\ref{table:detection_analysis_detailed} \\
		Table~\ref{table:avp} & Table~\ref{table:avp_detailed} \\
		\hline
	\end{tabular}
\caption{An overview of all tables in Appendix \ref{sec:more_results}.}
\label{table:list_of_tables}
\end{table}

\begin{table*}
	\centering
	\setlength{\tabcolsep}{1.3mm}
	\begin{tabular}{|c|c|cccccccccccc|c|}
		\hline
		Metric & Model & aero & bike & boat & bottle & bus & car & chair & dtable & mbike & sofa & train & tv & Mean \\
		\hline
		\multirow{2}{*}{$MedErr$} & $\R_E$ & 14.5 & 17.7 & 39.3 & 7.4 & 4.0 & 7.8 & 15.2 & 26.6 & 17.5 & \textbf{10.5} & 11.5 & 14.1 & 15.50 \\ 
		& $\R_G$ & \textbf{11.8} & \textbf{15.9} & \textbf{27.2} & \textbf{7.2} & \textbf{2.9} & \textbf{5.2} & \textbf{11.6} & \textbf{15.0} & \textbf{14.3} & 10.8 & \textbf{5.4} & \textbf{12.4} & \textbf{11.63} \\ 
		\hline
		\multirow{2}{*}{$Acc_\frac{\pi}{6}$} & $\R_E$ & 0.77 & 0.75 & 0.41 & 0.96 & 0.91 & 0.83 & 0.72 & 0.56 & 0.75 & 0.90 & 0.75 & \textbf{0.87} & 0.7656 \\ 
		& $\R_G$ & \textbf{0.80} & \textbf{0.78} & \textbf{0.54} & \textbf{0.97} & \textbf{0.95} & \textbf{0.93} & \textbf{0.83} & \textbf{0.59} & \textbf{0.82} & \textbf{0.91} & \textbf{0.81} & 0.86 & \textbf{0.8166} \\ 
		\hline
	\end{tabular}
	\caption{Comparison between the Euclidean and Geodesic regression models under two metrics. Lower is better for the $MedErr$ metric and higher is better for the $Acc_\frac{\pi}{6}$ metric. Best results are in bold.}
	\label{table:regression_detailed}
\end{table*}

\begin{table*}
	\centering
	\setlength{\tabcolsep}{1.5mm}
	\begin{tabular}{|c|c|cccccccccccc|c|}
		\hline
		Metric & Dict. Size & aero & bike & boat & bottle & bus & car & chair & dtable & mbike & sofa & train & tv & Mean \\
		\hline
		\multirow{4}{*}{$MedErr$} & 50 & 14.9 & 16.6 & 24.5 & 10.8 & 6.1 & 8.6 & 15.4 & 41.1 & 15.0 & 12.5 & 8.3 & 14.4 & 15.68 \\ 
		& 100 & 13.3 & \textbf{16.1} & 24.3 & 9.1 & 4.5 & 7.0 & 12.8 & 31.3 & 14.0 & 11.0 & 7.2 & 12.9 & 13.64 \\ 
		& 200 & 10.7 & 16.5 & \textbf{23.7} & 8.2 & 4.0 & 6.2 & 12.8 & \textbf{25.1} & \textbf{13.2} & \textbf{10.0} & 7.1 & \textbf{12.5} & \textbf{12.50} \\ 
		& 400 & \textbf{10.3} & 16.9 & 24.6 & \textbf{7.6} & \textbf{3.7} & \textbf{5.3} & \textbf{12.6} & 29.8 & 13.6 & 10.4 & \textbf{6.0} & 13.3 & 12.85 \\ 
		\hline
		\multirow{4}{*}{$Acc_\frac{\pi}{6}$} & 50 & 0.84 & 0.76 & 0.57 & \textbf{0.93} & 0.98 & 0.91 & 0.77 & 0.50 & \textbf{0.86} & \textbf{0.91} & 0.86 & 0.85 & 0.8127 \\ 
		& 100 & 0.86 & \textbf{0.78} & 0.56 & 0.91 & \textbf{0.99} & \textbf{0.93} & 0.78 & 0.50 & 0.85 & 0.89 & \textbf{0.87} & \textbf{0.89} & 0.8170 \\ 
		& 200 & \textbf{0.87} & 0.77 & \textbf{0.58} & 0.92 & \textbf{0.99} & 0.92 & 0.78 & \textbf{0.54} & \textbf{0.86} & 0.88 & 0.84 & \textbf{0.89} & \textbf{0.8206} \\ 
		& 400 & 0.86 & 0.76 & 0.56 & \textbf{0.93} & 0.97 & 0.91 & \textbf{0.81} & 0.53 & 0.82 & \textbf{0.91} & 0.85 & \textbf{0.89} & 0.8162 \\ 
		\hline
	\end{tabular}
	\caption{Performance of the pure classification models on the validation set under two metrics for different dictionary sizes. Lower is better for the MedErr metric and higher is better for the Accuracy metric. Best results are in bold.}
	\label{table:hyperparameterK_detailed}
\end{table*}

%\begin{table*}
%	\centering
%	\setlength{\tabcolsep}{1.5mm}
%	\begin{tabular}{|c|c|cccccccccccc|c|}
%		\hline
%		Metric & Dict. Size & aero & bike & boat & bottle & bus & car & chair & dtable & mbike & sofa & train & tv & Mean \\
%		\hline
%		\multirow{2}{*}{$MedErr$} & 100 & 11.7 & 15.3 & 21.5 & 9.3 & 4.1 & 7.4 & 11.2 & 17.8 & 17.0 & 11.0 & 7.0 & 13.1 & 12.20 \\ 
%		& 200 & 11.3 & 15.8 & 21.0 & 8.3 & 4.0 & 6.1 & 10.5 & 12.3 & 16.4 & 10.6 & 6.6 & 12.8 & 11.31 \\ 
%		\hline
%		\multirow{2}{*}{$Acc_\frac{\pi}{6}$} & 100 & 0.84 & 0.77 & 0.60 & 0.95 & 0.97 & 0.95 & 0.90 & 0.63 & 0.78 & 0.94 & 0.81 & 0.87 & 0.8350 \\ 
%		& 200 & 0.82 & 0.75 & 0.59 & 0.95 & 0.97 & 0.93 & 0.90 & 0.62 & 0.79 & 0.96 & 0.82 & 0.86 & 0.8298 \\ 
%		\hline
%	\end{tabular}
%	\caption{Performance of the pure classification models on the test set. Lower is better for the MedErr metric and higher is better for the Accuracy metric.}
%	\label{table:classification_detailed}
%\end{table*}

\begin{table*}
	\centering
	\setlength{\tabcolsep}{1.5mm}
	\begin{tabular}{|c|c|cccccccccccc|c|}
		\hline
		Metric & Model & aero & bike & boat & bottle & bus & car & chair & dtable & mbike & sofa & train & tv & Mean \\
		\hline
		$MedErr$ & $\mathcal{C}$ & 11.3 & 15.8 & 21.0 & 8.3 & 4.0 & 6.1 & 10.5 & 12.3 & 16.4 & 10.6 & 6.6 & 12.8 & 11.31 \\ 
		\hline
		$Acc_\frac{\pi}{6}$ & $\mathcal{C}$ & 0.82 & 0.75 & 0.59 & 0.95 & 0.97 & 0.93 & 0.90 & 0.62 & 0.79 & 0.96 & 0.82 & 0.86 & 0.8298 \\ 
		\hline
	\end{tabular}
	\caption{Performance of the pure classification model on the test set with dictionary size $K=200$. Lower is better for the MedErr metric and higher is better for the Accuracy metric.}
	\label{table:classification_detailed}
\end{table*}

\begin{table*}
	\centering
	\setlength{\tabcolsep}{1.5mm}
	\begin{tabular}{|c|c|cccccccccccc|c|}
		\hline
		Metric & $\alpha$ & aero & bike & boat & bottle & bus & car & chair & dtable & mbike & sofa & train & tv & Mean \\
		\hline
		\multirow{3}{*}{$MedErr$} & 0.1 & \textbf{11.0} & 16.1 & 23.8 & 8.7 & \textbf{3.4} & 5.8 & 12.2 & 30.1 & 13.6 & \textbf{9.1} & 6.8 & 13.3 & 12.83 \\ 
		& 1.0 & \textbf{11.0} & \textbf{15.9} & \textbf{22.7} & \textbf{8.2} & 3.6 & 6.0 & \textbf{12.1} & \textbf{21.0} & \textbf{13.4} & 9.7 & \textbf{6.2} & 13.2 & \textbf{11.92} \\ 
		& 10.0 & 11.4 & 16.6 & 23.9 & 8.3 & \textbf{3.4} & \textbf{5.6} & 12.4 & 40.3 & 13.8 & 11.0 & 6.7 & \textbf{13.1} & 13.90 \\ 
		\hline
		\multirow{3}{*}{$Acc_\frac{\pi}{6}$} & 0.1 & 0.86 & \textbf{0.77} & \textbf{0.57} & 0.92 & 0.97 & 0.92 & 0.78 & 0.51 & \textbf{0.83} & 0.89 & \textbf{0.87} & 0.88 & 0.8145 \\ 
		& 1.0 & \textbf{0.87} & 0.75 & \textbf{0.57} & \textbf{0.93} & 0.97 & 0.93 & \textbf{0.80} & \textbf{0.55} & \textbf{0.83} & 0.91 & 0.86 & \textbf{0.89} & \textbf{0.8212} \\ 
		& 10.0 & 0.86 & \textbf{0.77} & 0.55 & \textbf{0.93} & 0.97 & 0.93 & 0.78 & 0.44 & 0.80 & \textbf{0.93} & 0.85 & 0.88 & 0.8068 \\ 
		\hline
	\end{tabular}
	\caption{Performance of the Geodesic Bin \& Delta model $\M_G$ with dictionary size 200 on the validation set for different choices of weighting parameter $\alpha$. Lower is better for the MedErr metric and higher is better for the Accuracy metric.}
	\label{table:hyperparameteralpha_detailed}
\end{table*}

\begin{table*}
	\centering
	\setlength{\tabcolsep}{1.5mm}
	\begin{tabular}{|c|l|cccccccccccc|c|}
		\hline
		Metric & Model & aero & bike & boat & bottle & bus & car & chair & dtable & mbike & sofa & train & tv & Mean \\
		\hline
		\multirow{13}{*}{$MedErr$} & $\mathcal{M}_S$ & 11.0 & 15.5 & 21.0 & 8.8 & 3.8 & 7.0 & 10.8 & 21.0 & 16.6 & 10.7 & 6.5 & 13.1 & 12.14 \\ 
		& $\mathcal{M}_S+$ & 12.2 & 15.7 & 24.4 & 9.9 & 3.6 & 6.5 & 12.0 & 14.8 & 14.4 & 11.9 & 6.4 & 11.6 & 11.95 \\ 
		& $\mathcal{M}_G$ & 10.6 & 16.4 & 21.6 & 8.1 & 3.2 & 6.0 & 9.9 & 14.6 & 16.0 & 11.1 & 6.3 & 13.4 & 11.44 \\ 
		& $\mathcal{M}_G+$ & \textbf{8.5} & \textbf{14.8} & \textbf{20.5} & 7.0 & 3.1 & 5.1 & \textbf{9.3} & \textbf{11.3} & 14.2 & 10.2 & 5.6 & \textcolor{red}{11.7} & \textbf{10.10} \\ 
		& $\mathcal{M}_R$ & 11.3 & 16.2 & 21.6 & 8.4 & 3.4 & 6.0 & 10.6 & 16.8 & 16.2 & 12.1 & 5.9 & 12.0 & 11.69 \\ 
		& $\mathcal{M}_R+$ & \textcolor{red}{9.1} & 15.3 & \textcolor{red}{20.9} & 7.3 & 2.9 & \textcolor{red}{5.0} & \textcolor{red}{9.5} & 15.1 & 14.5 & \textbf{9.4} & 5.7 & \textbf{11.5} & \textcolor{red}{10.52} \\ 
		& $\mathcal{M}_{LE}$ & 12.8 & 15.2 & 23.4 & 9.0 & 4.0 & 7.4 & 11.1 & 16.8 & 16.1 & 10.7 & 6.6 & 12.3 & 12.11 \\ 
		& $\mathcal{M}_{LE}+$ & 12.3 & 16.7 & 24.7 & 7.5 & 3.6 & 6.5 & 11.5 & 15.5 & 15.1 & 11.1 & 7.3 & 12.1 & 11.99 \\ 
		& $\mathcal{M}_P$ & 10.7 & 15.9 & 21.3 & 8.3 & 3.4 & 5.8 & 10.4 & 14.5 & 16.0 & 12.4 & 6.7 & 12.6 & 11.50 \\ 
		& $\mathcal{M}_P+$ & 9.8 & \textcolor{red}{14.9} & 23.6 & \textcolor{red}{6.8} & \textbf{2.5} & \textbf{4.7} & 10.1 & 14.3 & \textcolor{red}{13.8} & 11.7 & 5.6 & 11.8 & 10.80 \\ 		
		& $\mathcal{M}_X+$ & 10.8 & 15.3 & 23.3 & 7.3 & 3.3 & 5.9 & 10.7 & 17.0 & 15.1 & 10.6 & 6.3 & 12.9 & 11.53 \\ 
		& $\mathcal{M}_{XP}$ & 11.4 & 16.3 & 25.6 & 7.0 & \textcolor{red}{2.6} & 5.1 & 11.3 & 16.0 & \textbf{13.6} & 10.2 & \textcolor{red}{5.5} & 12.0 & 11.38 \\ 
		& $\mathcal{M}_{XP}+$ & 10.6 & 15.0 & 23.9 & \textbf{6.7} & 2.7 & \textbf{4.7} & 9.8 & \textcolor{red}{12.6} & 13.9 & \textcolor{red}{9.7} & \textbf{5.3} & \textcolor{red}{11.7} & \textcolor{red}{10.54} \\ 		
		\hline
		\multirow{13}{*}{$Acc_\frac{\pi}{6}$} & $\mathcal{M}_S$ & 0.83 & 0.78 & 0.61 & \textcolor{red}{0.96} & 0.96 & 0.94 & 0.90 & 0.56 & 0.79 & 0.95 & \textcolor{red}{0.82} & 0.87 & 0.8303 \\
		& $\mathcal{M}_S+$ & 0.82 & 0.80 & 0.59 & 0.94 & 0.97 & 0.94 & 0.91 & 0.63 & 0.81 & \textbf{0.97} & \textbf{0.83} & 0.87 & 0.8387 \\
		& $\mathcal{M}_G$ & 0.84 & 0.76 & \textcolor{red}{0.62} & \textcolor{red}{0.96} & \textbf{0.98} & 0.94 & \textbf{0.92} & 0.65 & 0.80 & \textcolor{red}{0.96} & \textcolor{red}{0.82} & 0.87 & 0.8439 \\ 
		& $\mathcal{M}_G+$ & \textbf{0.87} & \textcolor{red}{0.81} & \textbf{0.64} & \textcolor{red}{0.96} & \textcolor{red}{0.97} & \textcolor{red}{0.95} & \textbf{0.92} & 0.67 & \textbf{0.85} & \textbf{0.97} & \textcolor{red}{0.82} & 0.88 & \textbf{0.8588} \\ 
		& $\mathcal{M}_R$ & 0.85 & 0.77 & 0.60 & 0.95 & \textcolor{red}{0.97} & 0.94 & \textbf{0.92} & 0.54 & 0.80 & 0.93 & 0.81 & 0.86 & 0.8285 \\ 
		& $\mathcal{M}_R+$ & \textcolor{red}{0.86} & \textcolor{red}{0.81} & \textcolor{red}{0.62} & \textcolor{red}{0.96} & \textcolor{red}{0.97} & \textcolor{red}{0.95} & \textbf{0.92} & 0.67 & 0.83 & \textbf{0.97} & \textbf{0.83} & \textcolor{red}{0.90} & \textcolor{red}{0.8573} \\ 
		& $\mathcal{M}_{LE}$ & 0.83 & 0.77 & 0.58 & \textcolor{red}{0.96} & 0.96 & 0.94 & \textcolor{red}{0.91} & \textcolor{red}{0.71} & 0.81 & 0.93 & 0.81 & 0.87 & 0.8410 \\ 
		& $\mathcal{M}_{LE}+$ & 0.81 & 0.77 & 0.56 & \textcolor{red}{0.96} & \textcolor{red}{0.97} & 0.92 & 0.86 & \textbf{0.73} & 0.79 & 0.93 & 0.80 & 0.89 & 0.8329 \\ 
		& $\mathcal{M}_P$ & 0.85 & 0.74 & \textcolor{red}{0.62} & 0.95 & \textbf{0.98} & \textbf{0.96} & \textbf{0.92} & \textcolor{red}{0.71} & 0.82 & 0.93 & \textcolor{red}{0.82} & \textbf{0.91} & 0.8491 \\ 
		& $\mathcal{M}_P+$ & 0.85 & 0.79 & 0.60 & \textcolor{red}{0.96} & \textcolor{red}{0.97} & \textcolor{red}{0.95} & 0.88 & 0.68 & 0.82 & 0.93 & 0.81 & 0.89 & 0.8457 \\ 
		& $\mathcal{M}_X+$ & 0.83 & \textcolor{red}{0.81} & 0.61 & \textbf{0.97} & 0.96 & 0.94 & 0.86 & 0.65 & 0.80 & 0.95 & 0.81 & 0.89 & 0.8407 \\ 
		& $\mathcal{M}_{XP}$ & 0.80 & 0.77 & 0.56 & \textbf{0.97} & \textcolor{red}{0.97} & 0.93 & 0.82 & 0.57 & 0.81 & 0.92 & \textcolor{red}{0.82} & 0.88 & 0.8185 \\ 
		& $\mathcal{M}_{XP}+$ & 0.84 & \textbf{0.82} & 0.59 & \textbf{0.97} & \textcolor{red}{0.97} & \textcolor{red}{0.95} & 0.88 & 0.68 & \textcolor{red}{0.84} & 0.93 & 0.81 & 0.89 & 0.8470 \\ 		
		\hline
	\end{tabular}
	\caption{Performance of the Bin \& Delta models on the test set under different metrics. Lower is better for the MedErr metric and higher is better for the Accuracy metric. Best results are highlighted in bold and second-best results are shown in red (best seen in color).}
	\label{table:all_bd_detailed}
\end{table*}

\begin{table*}
	\centering
	\setlength{\tabcolsep}{1.5mm}
	\begin{tabular}{|c|cc|cccccccccccc|c|}
		\hline
		Metric & K & $\gamma$ & aero & bike & boat & bottle & bus & car & chair & dtable & mbike & sofa & train & tv & Mean \\
		\hline
		\multirow{4}{*}{$MedErr$} & 16 & 2.06 & 12.7 & \textbf{16.9} & \textbf{24.4} & \textbf{8.0} & \textbf{3.1} & \textbf{5.7} & \textbf{13.2} & \textbf{16.0} & \textbf{16.3} & \textbf{12.6} & 6.7 & 14.1 & \textbf{12.48} \\ 
		& 50 & 7.82 & \textbf{12.4} & 17.4 & 26.3 & 9.0 & \textbf{3.1} & 6.1 & 13.4 & 16.6 & 17.7 & 13.7 & \textbf{6.4} & \textbf{13.7} & 12.98 \\ 
		& 100 & 15.08 & 86.0 & 81.8 & 75.6 & 11.6 & 23.8 & 77.3 & 84.6 & 23.6 & 70.4 & 45.3 & 17.0 & 23.4 & 51.71 \\
		& 200 & 25.23 & 75.0 & 87.0 & 81.2 & 11.0 & 19.2 & 46.9 & 80.2 & 22.2 & 58.5 & 43.0 & 13.1 & 23.4 & 46.73 \\ 
		\hline
		\multirow{4}{*}{$Acc_\frac{\pi}{6}$} & 16 & 2.06 & \textbf{0.83} & \textbf{0.81} & \textbf{0.62} & \textbf{0.95} & \textbf{0.96} & 0.93 & \textbf{0.87} & 0.62 & \textbf{0.84} & \textbf{0.90} & \textbf{0.81} & 0.83 & 0.8302 \\ 
		& 50 & 7.82 & 0.82 & 0.79 & 0.56 & \textbf{0.95} & 0.95 & \textbf{0.95} & 0.86 & \textbf{0.71} & 0.80 & \textbf{0.90} & \textbf{0.81} & \textbf{0.87} & \textbf{0.8320} \\ 
		& 100 & 15.08 & 0.15 & 0.14 & 0.17 & 0.86 & 0.59 & 0.26 & 0.16 & 0.52 & 0.24 & 0.28 & 0.69 & 0.65 & 0.3932 \\ 
		& 200 & 25.23 & 0.24 & 0.13 & 0.18 & 0.87 & 0.63 & 0.36 & 0.20 & 0.57 & 0.21 & 0.33 & 0.72 & 0.64 & 0.4236 \\ 
		\hline
	\end{tabular}
	\caption{Performance of the RelaXed Bin \& Delta models $\M_X$ on the validation set under two metrics for different dictionary sizes. Lower is better for the MedErr metric and higher is better for the Accuracy metric.}
	\label{table:hyperparametergamma_detailed}
\end{table*}

\begin{table*}
	\centering
	\setlength{\tabcolsep}{1.3mm}
	\begin{tabular}{|c|c|c|cccccccccccc|c|}
		\hline
		Metric & Model & Representation & aero & bike & boat & bottle & bus & car & chair & dtable & mbike & sofa & train & tv & Mean \\
		\hline
		\multirow{8}{*}{$MedErr$} & \multirow{2}{*}{$\mathcal{R}_E$} & Axis-angle & 14.5 & 17.7 & 39.3 & 7.4 & 4.0 & 7.8 & 15.2 & 26.6 & 17.5 & 10.5 & 11.5 & 14.1 & 15.50 \\ 
		& & Quaternion & 12.7 & 16.7 & 34.9 & 7.3 & 4.0 & 6.7 & 13.4 & 25.1 & 15.7 & \textbf{10.3} & 8.9 & 13.9 & 14.14 \\ 
		\cline{2-16}
		& \multirow{2}{*}{$\mathcal{R}_G$} & Axis-angle & \textbf{11.8} & \textbf{15.9} & \textbf{27.2} & \textbf{7.2} & \textbf{2.9} & \textbf{5.2} & \textbf{11.6} & \textbf{15.0} & \textbf{14.3} & 10.8 & \textbf{5.4} & \textbf{12.4} & \textbf{11.63} \\ 
		& & Quaternion & 12.2 & 16.0 & 29.8 & 7.5 & \textbf{2.9} & 5.8 & 12.1 & 16.8 & 14.8 & \textbf{10.3} & 5.9 & 12.4 & 12.20 \\ 
		\cline{2-16}
		& \multirow{2}{*}{$\mathcal{M}_G$} & Axis-angle & 10.6 & 16.4 & 21.6 & 8.1 & 3.2 & 6.0 & 9.9 & 14.6 & 16.0 & 11.1 & 6.3 & 13.4 & 11.44 \\ 
		& & Quaternion & 10.7 & 15.5 & 21.0 & 7.4 & 3.0 & 5.2 & 9.8 & 19.0 & 15.3 & 10.0 & 5.4 & 12.4 & 11.23 \\ 
		\cline{2-16}
		& \multirow{2}{*}{$\mathcal{M}_G+$} & Axis-angle & 8.5 & 14.8 & 20.5 & 7.0 & 3.1 & 5.1 & 9.3 & 11.3 & 14.2 & 10.2 & 5.6 & 11.7 & 10.10 \\ 
		& & Quaternion & 9.2 & 14.3 & 22.1 & 7.2 & 2.9 & 4.9 & 9.3 & 17.4 & 13.6 & 11.2 & 5.2 & 11.8 & 10.75 \\ 
		\hline
		\multirow{8}{*}{$Acc_\frac{\pi}{6}$} & \multirow{2}{*}{$\mathcal{R}_E$} & Axis-angle & 0.77 & 0.75 & 0.41 & 0.96 & 0.91 & 0.83 & 0.72 & 0.56 & 0.75 & 0.90 & 0.75 & \textbf{0.87} & 0.7656 \\ 
		& & Quaternion & 0.78 & \textbf{0.80} & 0.44 & \textbf{0.97} & 0.95 & 0.89 & 0.78 & 0.57 & 0.78 & \textbf{0.91} & 0.80 & \textbf{0.87} & 0.7965 \\ 
		\cline{2-16}
		& \multirow{2}{*}{$\mathcal{R}_G$} & Axis-angle & 0.80 & 0.78 & \textbf{0.54} & \textbf{0.97} & 0.95 & \textbf{0.93} & \textbf{0.83} & 0.59 & \textbf{0.82} & \textbf{0.91} & 0.81 & 0.86 & \textbf{0.8166} \\ 
		& & Quaternion & \textbf{0.81} & 0.79 & 0.51 & 0.96 & \textbf{0.97} & 0.91 & 0.82 & \textbf{0.60} & 0.81 & \textbf{0.91} & \textbf{0.82} & \textbf{0.87} & 0.8141 \\ 
		\cline{2-16}
		& \multirow{2}{*}{$\mathcal{M}_G$} & Axis-angle & 0.84 & 0.76 & 0.62 & 0.96 & 0.98 & 0.94 & 0.92 & 0.65 & 0.80 & 0.96 & 0.82 & 0.87 & 0.8439 \\ 
		& & Quaternion & 0.84 & 0.77 & 0.62 & 0.96 & 0.97 & 0.95 & 0.91 & 0.57 & 0.81 & 0.94 & 0.82 & 0.90 & 0.8384 \\ 
		\cline{2-16}
		& \multirow{2}{*}{$\mathcal{M}_G+$} & Axis-angle & 0.87 & 0.81 & 0.64 & 0.96 & 0.97 & 0.95 & 0.92 & 0.67 & 0.85 & 0.97 & 0.82 & 0.88 & 0.8588 \\ 
		& & Quaternion & 0.86 & 0.81 & 0.62 & 0.96 & 0.98 & 0.95 & 0.94 & 0.60 & 0.84 & 0.97 & 0.83 & 0.90 & 0.8560 \\ 
		\hline
	\end{tabular}
	\caption{Comparison between the Axis-angle and Quaternion representations for different 3D pose estimation models.}
	\label{table:representation_detailed}
\end{table*}

\begin{table*}
	\centering
	\setlength{\tabcolsep}{1.5mm}
	\begin{tabular}{|c|c|c|cccccccccccc|c|}
		\hline
		Metric & Model & Data & aero & bike & boat & bottle & bus & car & chair & dtable & mbike & sofa & train & tv & Mean \\
		\hline
		\multirow{6}{*}{$MedErr$} & \multirow{3}{*}{$\mathcal{M}_G$} & Rendered & 12.7 & 18.0 & 38.5 & 10.1 & 7.4 & 7.3 & 11.1 & 12.1 & 21.7 & 12.8 & 11.4 & 14.3 & 14.79 \\ 
		& & Augmented & 15.8 & 29.9 & 35.5 & 11.4 & 3.4 & 7.2 & 30.6 & 18.7 & 22.5 & 18.6 & 7.9 & 18.6 & 18.35 \\ 
		& & Both & 10.6 & 16.4 & 21.6 & 8.1 & 3.2 & 6.0 & 9.9 & 14.6 & 16.0 & 11.1 & 6.3 & 13.4 & 11.44 \\ 
		\cline{2-16}
		& \multirow{3}{*}{$\mathcal{M}_G+$} & Rendered & 12.0 & 18.6 & 38.3 & 9.1 & 8.3 & 6.9 & 10.6 & 21.2 & 19.9 & \textbf{9.8} & 11.6 & 14.0 & 15.03 \\ 
		& & Augmented & 12.2 & 21.9 & 27.0 & 8.9 & \textbf{2.8} & 5.3 & 14.6 & 25.3 & 17.5 & 16.7 & 6.1 & 13.4 & 14.30 \\ 
		& & Both & \textbf{8.5} & \textbf{14.8} & \textbf{20.5} & \textbf{7.0} & 3.1 & \textbf{5.1} & \textbf{9.3} & \textbf{11.3} & \textbf{14.2} & 10.2 & \textbf{5.6} & \textbf{11.7} & \textbf{10.10} \\ 
		\hline
		\multirow{6}{*}{$Acc_\frac{\pi}{6}$} & \multirow{3}{*}{$\mathcal{M}_G$} & Rendered & 0.78 & 0.76 & 0.42 & 0.95 & 0.84 & 0.90 & 0.87 & 0.68 & 0.66 & 0.91 & 0.71 & 0.84 & 0.7779 \\ 
		& & Augmented & 0.70 & 0.50 & 0.46 & 0.85 & 0.89 & 0.82 & 0.50 & 0.56 & 0.60 & 0.72 & 0.79 & 0.71 & 0.6757 \\ 
		& & Both & 0.84 & 0.76 & 0.62 & \textbf{0.96} & \textbf{0.98} & 0.94 & \textbf{0.92} & 0.65 & 0.80 & 0.96 & \textbf{0.82} & 0.87 & 0.8439 \\ 
		\cline{2-16}
		& \multirow{3}{*}{$\mathcal{M}_G+$} & Rendered & 0.78 & 0.75 & 0.41 & \textbf{0.96} & 0.83 & 0.89 & 0.87 & 0.65 & 0.66 & 0.95 & 0.69 & 0.85 & 0.7748 \\ 
		& & Augmented & 0.77 & 0.63 & 0.54 & 0.94 & 0.97 & 0.90 & 0.76 & 0.54 & 0.71 & 0.66 & 0.79 & 0.80 & 0.7506 \\ 
		& & Both & \textbf{0.87} & \textbf{0.81} & \textbf{0.64} & \textbf{0.96} & 0.97 & \textbf{0.95} & \textbf{0.92} & \textbf{0.67} & \textbf{0.85} & \textbf{0.97} & \textbf{0.82} & \textbf{0.88} & \textbf{0.8588} \\ 
		\hline
	\end{tabular}
	\caption{Performance of the Geodesic Bin \& Delta models on the test set with different types of training data. Lower is better for the MedErr metric and higher is better for the Accuracy metric.}
	\label{table:data_augmentation_detailed}
\end{table*}

\begin{table*}
	\centering
	\setlength{\tabcolsep}{1.3mm}
	\begin{tabular}{|c|l|c|cccccccccccc|c|}
		\hline
		Metric & Model & Representation & aero & bike & boat & bottle & bus & car & chair & dtable & mbike & sofa & train & tv & Mean \\
		\hline
		\multirow{4}{*}{$MedErr$} & \multirow{2}{*}{$\mathcal{M}_G$} & ResNet50 & 10.6 & 16.4 & 21.6 & 8.1 & 3.2 & 6.0 & 9.9 & 14.6 & 16.0 & 11.1 & 6.3 & 13.4 & 11.44 \\ 
		& & VGG13 & 10.9 & 16.4 & 27.9 & 8.5 & 3.1 & 5.9 & 10.9 & \textbf{10.7} & 16.1 & 12.0 & 6.5 & 13.2 & 11.83 \\ 
		\cline{2-16}
		& \multirow{2}{*}{$\mathcal{M}_G+$} & ResNet50 & \textbf{8.5} & \textbf{14.8} & \textbf{20.5} & \textbf{7.0} & 3.1 & 5.1 & \textbf{9.3} & 11.3 & 14.2 & \textbf{10.2} & \textbf{5.6} & 11.7 & \textbf{10.10} \\ 
		& & VGG13 & 10.2 & 15.4 & 25.6 & 8.0 & \textbf{2.8} & \textbf{5.0} & 10.2 & 10.9 & \textbf{13.4} & 10.8 & 5.9 & \textbf{11.4} & 10.81 \\ 
		\hline
		\multirow{4}{*}{$Acc_\frac{\pi}{6}$} & \multirow{2}{*}{$\mathcal{M}_G$} & ResNet50 & 0.84 & 0.76 & 0.62 & \textbf{0.96} & \textbf{0.98} & 0.94 & 0.92 & 0.65 & 0.80 & 0.96 & 0.82 & 0.87 & 0.8439 \\ 
		& & VGG13 & 0.84 & 0.77 & 0.53 & 0.95 & 0.95 & 0.93 & 0.89 & 0.67 & 0.82 & 0.93 & 0.81 & 0.85 & 0.8277 \\ 
		\cline{2-16}
		& \multirow{2}{*}{$\mathcal{M}_G+$} & ResNet50 & \textbf{0.87} & 0.81 & \textbf{0.64} & \textbf{0.96} & 0.97 & \textbf{0.95} & \textbf{0.92} & 0.67 & \textbf{0.85} & \textbf{0.97} & 0.82 & \textbf{0.88} & \textbf{0.8588} \\ 
		& & VGG13 & 0.83 & \textbf{0.82} & 0.55 & \textbf{0.96} & \textbf{0.98} & 0.94 & 0.89 & \textbf{0.70} & 0.82 & 0.93 & \textbf{0.83} & \textbf{0.88} & 0.8423 \\
		\hline
	\end{tabular}
	\caption{Performance of the Geodesic Bin \& Delta models on the test set using different feature networks. Lower is better for the MedErr metric and higher is better for the Accuracy metric.}
	\label{table:feature_network_detailed}
\end{table*}

\begin{table*}
	\centering
	\setlength{\tabcolsep}{1mm}
	\begin{tabular}{|c|c|cccccccccccc|c|}
		\hline
		Model & Detections & aero & bike & boat & bottle & bus & car & chair & dtable & mbike & sofa & train & tv & Mean \\
		\hline
		\citep{Tulsiani:CVPR15} & \multirow{3}{*}{V\&K} & 64.0 & 53.2 & 21.0 & - & 69.3 & 55.1 & 24.6 & 16.9 & 54.0 & 42.5 & 59.4 & 51.2 & 46.5 \\
		$\M_G$ & & 67.6& 56.4 & 25.7 & - &	73.2 & 59.2 & 30.6 & 20.9 & 62.0 & 	50.5 & 64.6 & 57.1 & 51.6 \\
		$\M_G+$ & & 70.2 & 61.1 & 27.3 & - & 73.5 & \textbf{59.6} & 31.2 & 22.3 & 62.1	& 48.3 & 64.4 & 58.8 & 52.6 \\
		\hline
		$\M_G$ & \multirow{2}{*}{R4CNN} & 69.1 & 57.4 & 25.0 & - & 69.8 & 56.5 & 24.2 & 22.8 & 61.6 & 45.9 & 57.1 & 62.5 & 50.2 \\
		$\M_G+$ & & 72.1 & 60.6& 25.8 & - & 69.4 & 56.6 & 24.7 & 25.1 & 62.2 &	46.0 & 57.8 & 64.1 & 51.3 \\
		\hline
		$\M_G$ & \multirow{2}{*}{Mask-RCNN} & 78 & 62.1 & 39.9 & - & 84.2 & 58.0 & 49.2 & 31.6 & 74.4	& 53.7 & 75.7 & 70.3 & 61.6 \\
		$\M_G+$ & & \textbf{81.3} & \textbf{68.9} & \textbf{42.7} & - & \textbf{85.5} & 58.8 & \textbf{50.3} & \textbf{36.1} & \textbf{76.3} & \textbf{55.0} & \textbf{76.8} &	\textbf{71.3} & \textbf{63.9} \\
		\hline
	\end{tabular}
	\caption{Performance of our Geodesic Bin \& Delta models under the ARP metric using detected bounding boxes. We use the bounding boxes provided by V\&K \citep{Tulsiani:CVPR15} and R4CNN  \citep{Su:ICCV15}. Higher is better.}
	\label{table:arp_detailed}
\end{table*}

\begin{table*}
	\centering
	\begin{tabular}{|c|cccccccccccc|c|}
		\hline
		Detections & aero & bike & boat & bottle & bus & car & chair & dtable & mbike & sofa & train & tv & Mean \\
		\hline
		V\&K & 75.1 & 72.6 & 38.8 & 41.7 & 76.1 & 65.1 & 36.3 & 41.6 & 75.7 & 58.8 & 75.4 & 61.8 & 61.6 \\
		R4CNN & 76.6 & 72.9 & 37.4 & 38.4 & 71.8 & 61.6 & 31.1 & 46.9 & 77.4 & 58.7 & 67.5 & 69.0 &	61.0 \\
		Mask-RCNN & \textbf{89.0} & \textbf{81.6} & \textbf{63.5} & \textbf{75.3} & \textbf{89.1} & \textbf{71.5} & \textbf{62.6} & \textbf{60.7} & \textbf{89.3} & \textbf{64.8} & \textbf{90.7} & \textbf{75.8} & \textbf{76.2} \\
		\hline
	\end{tabular}
	\caption{Performance of the detected bounding boxes under the AP metric. Higher is better.}
	\label{table:ap_detailed}
\end{table*}

\begin{table*}
	\centering
	\setlength{\tabcolsep}{1.2mm}
	\begin{tabular}{|c|c|c|cccccccccccc|c|}
		\hline
		Model & Detection & Metric & aero & bike & boat & bottle & bus & car & chair & dtable & mbike & sofa & train & tv & Mean \\
		\hline
		& GT & \# BBoxes & 433	& 358 & 424	& 630 & 301	& 1004 & 1176 & 305 & 356 & 285	& 315 & 392 & \\
		\hline
		\multirow{9}{*}{$\M_G$} & \multirow{3}{*}{V\&K} & \% Detected & 0.85 & 0.83 & 0.56 & 0.56	& 0.80 & 0.73 & 0.50	& 0.60 & 0.88 & 0.87 & 0.91 & 0.78	& 0.7392 \\
		& & \% Correct & 0.66	& 0.54 & 0.31 & 0.51 & 0.69	& 0.57 & 0.34 & 0.30 & 0.59	& \textbf{0.58} & 0.69 &	0.64 & 0.5347 \\
		& & Pose-Err & 12.7	& 20.3 & 23.4 & 9.4	& 4.7 & 9.8	& 17.0 & 30.5 & 20.5 & 19.8	& 9.4 & 15.3	& 16.06 \\
		\cline{2-16}
		& \multirow{3}{*}{R4CNN} & \% Detected & 0.84 & 0.81 & 0.59 & 0.51 & 0.79 & 0.70 & 0.48 & 0.69	& 0.84 & 0.86 & 0.80	& 0.80 &	0.7258 \\
		& & \% Correct & 0.65	& 0.53 & 0.34 & 0.47 & 0.69	& 0.55 & 0.32 & 0.33 & 0.57	& 0.57 & 0.62 & 0.67 & 0.5264 \\
		& & Pose-Err & 12.3	& 19.3 & 23.9 & 9.0 & 4.5 & 9.6	& 17.6 & 33.5 & 19.6 & 19.7	& 9.0 & 14.3 & 16.01 \\
		\cline{2-16}
		& \multirow{3}{*}{Mask-RCNN} & \% Detected & \textbf{0.90}	& \textbf{0.84}	& \textbf{0.75}	& \textbf{0.79}	& \textbf{0.91}	& \textbf{0.80}	& \textbf{0.73}	& \textbf{0.72}	& \textbf{0.91}	& \textbf{0.75}	& \textbf{0.92}	& \textbf{0.77}	& \textbf{0.8158} \\
		& & \% Correct & 0.72	& 0.56	& 0.41	& 0.74	& \textbf{0.78}	& 0.61	& 0.49	& 0.39	& 0.63	& 0.53	& 0.69	& 0.68	& 0.6028 \\
		& & Pose-Err & 11.9	& 18.9	& 25.6	& 9.0	& 4.8	& 10.1	& 17.8	& 26.7	& 19.1	& 16.6	& 9.3	& 13.5	& 15.28 \\
		\hline
		\multirow{9}{*}{$\M_G+$} & \multirow{3}{*}{V\&K} & \% Detected & 0.85 & 0.83 & 0.56 & 0.56	& 0.80 & 0.73 & 0.50	& 0.60 & 0.88 & 0.87 & 0.91 & 0.78	& 0.7392 \\
		& & \% Correct & 0.69	& 0.58	& 0.32	& 0.52	& 0.69	& 0.58	& 0.35	& 0.32	& 0.61	& 0.57	& 0.68	& 0.66	& 0.5469 \\
		& & Pose-Err & 10.6	& 19.0	& \textbf{23.3}	& 8.9	& \textbf{4.3}	& 9.3	& \textbf{15.9}	& 27.8	& 18.6	& 18.9	& 8.9	& 14.2	& 14.96 \\
		\cline{2-16}
		& \multirow{3}{*}{R4CNN} & \% Detected & 0.84 & 0.81 & 0.59 & 0.51 & 0.79 & 0.70 & 0.48 & 0.69	& 0.84 & 0.86 & 0.80	& 0.80 &	0.7258 \\
		& & \% Correct & 0.70	& 0.56	& 0.34	& 0.48	& 0.68	& 0.55	& 0.33	& 0.36	& 0.59	& \textbf{0.58}	& 0.62	& \textbf{0.69}	& 0.5406\\
		& & Pose-Err & 10.0	& 19.4	& 24.2	& 8.4	& 4.4	& \textbf{9.2}	& 16.4	& 28.2	& 18.6	& 17.7	& \textbf{7.7}	& 12.9	& 14.77 \\
		\cline{2-16}
		& \multirow{3}{*}{Mask-RCNN} & \% Detected & \textbf{0.90}	& \textbf{0.84}	& \textbf{0.75}	& \textbf{0.79}	& \textbf{0.91}	& \textbf{0.80}	& \textbf{0.73}	& \textbf{0.72}	& \textbf{0.91}	& \textbf{0.75}	& \textbf{0.92}	& \textbf{0.77}	& \textbf{0.8158} \\
		& & \% Correct & \textbf{0.76}	& \textbf{0.61}	& \textbf{0.44}	& \textbf{0.75}	& \textbf{0.78}	& \textbf{0.62}	& \textbf{0.51}	& \textbf{0.41}	& \textbf{0.66}	& 0.54	& \textbf{0.70}	& 0.68	& \textbf{0.6225}\\
		& & Pose-Err & \textbf{9.8}	& \textbf{18.0}	& 23.5	& \textbf{8.0}	& \textbf{4.3}	& 9.5	& 16.1	& \textbf{23.6}	& \textbf{17.2}	& \textbf{15.9}	& 8.5	& \textbf{11.7}	& \textbf{13.83} \\
		\hline
	\end{tabular}
	\caption{Performance of the Geodesic Bin \& Delta models with detected bounding boxes. We report three metrics for each combination of model and detection system (i) \% Detected: the percentage of bounding boxes detected correctly by the detection system (IoU overlap $>0.5$), (ii) \% Correct: the percentage of bounding boxes detected correctly and having pose error $<30^\circ$, and (iii) MedErr: median pose error (in degrees) for all detected bounding boxes. Higher is better for the percentages and lower is better for Pose-Err.}
	\label{table:detection_analysis_detailed}
\end{table*}

\begin{table*}
	\centering
	\setlength{\tabcolsep}{1.2mm}
	\begin{tabular}{|c|cc|cccccccccccc|c|}
		\hline
		\# Bins & Model & Detections & aero & bike & boat & bottle & bus & car & chair & dtable & mbike & sofa & train & tv & Mean \\
		\hline
		\multirow{9}{*}{4} & V\&K & V\&K & 63.1 & 59.4 & 23.0 & - & 69.8 & 55.2 & 25.1 & 24.3 & 61.1 & 43.8 & 59.4 & 55.4 & 49.1 \\
		& R4CNN & R4CNN & 54.0 & 50.5 & 15.1 & - & 57.1 & 41.8 & 15.7 & 18.6 & 50.8 & 28.4 & 46.1 & 58.2 & 39.7 \\
		& Massa & - & 70.3 & 67.0 & 36.7 & - & 75.4 & \textbf{58.3} & 21.4 & 34.5 & 71.5 & 46.0 & 64.3 & 63.4 & 55.4 \\
		\cline{2-16}
		& \multirow{3}{*}{$\M_G$} & V\&K & 59.7 & 61.3 & 22.0 & - & 65.0 & 53.8 & 28.2 & 23.1 & 58.8 & 49.1 & 60.8 & 57.2 & 49.0 \\
		& & R4CNN & 60.4 & 60.8 & 22.3 & - & 63.4 & 50.2 & 23.4 & 26.7 & 59.3 & 46.7 & 53.7 & 63.2 & 48.2 \\
		& & Mask-RCNN & 70.2	& 66.9	& 37.0	& -	& 75.4	& 51.4	& 48.1	& 39.3	& 68.2	& 53	& 72.1	& 69.3	& 59.2 \\
		\cline{2-16}
		& \multirow{3}{*}{$\M_G+$} & V\&K & 61.0 & 59.7 & 24.7 & - & 68.1 & 55.7 & 29.3 & 22.9 &	61.2 &	47.6 &	61.6 &	57.1 & 49.9 \\
		& & R4CNN & 63.4 & 61.2 & 23.4 & - & 63.6 & 53.5 & 24.6 &	25.3 & 63.2 & 47.1 & 54.5 & 63.5 & 49.4 \\
		& & Mask-RCNN & \textbf{74.9}	& \textbf{69.8}	& \textbf{41.2}	& -	& \textbf{77.6}	& 55.9	& \textbf{50.1}	& \textbf{40.5}	& \textbf{74.1}	& \textbf{55.3}	& \textbf{72.2}	& \textbf{70.1}	& \textbf{62.0} \\
		\hline
		\multirow{9}{*}{8} & V\&K & V\&K &  57.5 & 54.8 & 18.9 & - & 59.4 & 51.5 & 24.7 & 20.4 & 59.5 & 43.7 & 53.3 & 45.6 & 44.5 \\
		& R4CNN & R4CNN & 44.5 & 41.1 & 10.1 & - & 48.0 & 36.6 & 13.7 & 15.1 & 39.9 & 26.8 & 39.1 & 46.5 & 32.9 \\
		& Massa & - & 66.0 & \textbf{62.5} & 31.2 & - & \textbf{68.7} & \textbf{55.7} & 19.2 & 31.9 & 64.0 & 44.7 & 61.8 & 58.0 & 51.3 \\
		\cline{2-16}
		& \multirow{3}{*}{$\M_G$} & V\&K & 47.8 & 51.5 & 17.1 & - & 52.2 & 46.6 & 25.7 & 22.4 & 53.2 & 42.7 & 51.1 & 45.9 & 41.5 \\
		& & R4CNN & 49.9 & 50.7 & 14.8 & - & 50.2 & 43.6 & 20.2 & 24.2 & 54.9 & 37.9 & 45.2 & 51.1 & 40.2 \\
		& & Mask-RCNN & 58.3	& 59.0	& 30.2	& -	& 60.0	& 42.5	& 42.7	& \textbf{35.2}	& 64.0	& 46.8	& 60.7	& 57.4	& 50.6\\
		\cline{2-16}
		& \multirow{3}{*}{$\M_G+$} & V\&K & 54.1 & 50.6 & 21.0 & - & 55.0 & 50.0 & 26.8 &	20.5 & 	55.3 & 42.0 & 51.9 & 47.2 &	43.1 \\
		& & R4CNN & 57.7 & 51.7 & 18.7 & - & 50.6 & 48.5 & 21.1 & 	24.0 & 54.8 & 39.3 & 47.5 & 53.4 & 42.5 \\
		& & Mask-RCNN & \textbf{66.7}	& 60.8	& \textbf{34.1}	& -	& 62.8	& 48.1	& \textbf{43.0}	& 32.6	& \textbf{67.3}	& \textbf{47.6}	& \textbf{63.2}	& \textbf{58.7}	& \textbf{53.2} \\
		\hline
		\multirow{9}{*}{16} & V\&K & V\&K & 46.6 & 42 & 12.7 & - & 64.6 & 42.8 & 20.8 & 18.5 & 38.8 & 33.5 & 42.4 & 32.9 & 36.0 \\
		& R4CNN & R4CNN & 27.5 & 25.8 & 6.5 & - & 45.8 & 29.7 & 8.5 & 12.0 & 31.4 & 17.7 & 29.7 & 31.4 & 24.2 \\
		& Massa & - & \textbf{51.4} & \textbf{43.0} & \textbf{23.6} & - & \textbf{68.9} & \textbf{46.3} & 15.2 & \textbf{29.3} & \textbf{49.4} & \textbf{35.6} & \textbf{47.0} & 37.3 & \textbf{40.6} \\
		\cline{2-16}
		& \multirow{3}{*}{$\M_G$} & V\&K & 37.8 & 35.3 & 11.0 & - & 48.4 & 37.6 & 21.8 & 157 & 33.0 & 33.7 & 39.8 & 32.0 & 31.5 \\
		& & R4CNN & 35.9 & 32.6 & 10.5 & - & 48.2 & 35.3 & 16.0 & 18.2 & 31.9 & 27.7 & 30.5 & 34.6 & 29.2 \\
		& & Mask-RCNN & 44.0	& 38.1	& 16.5	& - &	53.5	& 33.7	& 33.1	& 25.3	& 37.6	& 33.3	& 45.0	& 40.8	& 36.4 \\
		\cline{2-16}
		& \multirow{3}{*}{$\M_G+$} & V\&K & 43.3 & 34.7 & 13.5 & - & 50.4 & 40.5 & 23.2 & 	15.1 & 36.9 & 33.0 & 41.1 & 33.0 & 33.1 \\
		& & R4CNN & 44.2 & 35.1 & 12.2 & - & 47.1 & 40.4 & 16.9 & 16.4	& 38.6 & 28.8 & 33.2 & 38.6 & 32.0 \\
		& & Mask-RCNN & 51.1	& 41.7	& 19.0	& - &	57.6	& 37.3	& \textbf{35.5}	& 25.8	& 47.6	& 36.7	& 48.0	& \textbf{41.5}	& 40.2 \\
		\hline
		\multirow{9}{*}{24} & V\&K & V\&K & 37.0 & 33.4 & 10.0 & - & 54.1 & 40.0 & 17.5 & 19.9 & 34.3 & 28.9 & 43.9 & 22.7 & 31.1 \\
		& R4CNN & R4CNN & 21.5 & 22.0 & 4.1 & - & 38.6 & 25.5 & 7.4 & 11.0 & 24.4 & 15.0 & 28.0 & 19.8 & 19.8 \\
		& Massa & - & \textbf{43.2} & \textbf{39.4} & \textbf{16.8} & - & \textbf{61.0} & \textbf{44.2} & 13.5 & \textbf{29.4} & 37.5 & \textbf{33.5} & \textbf{46.6} & 32.5 & \textbf{36.1} \\
		\cline{2-16}
		&\multirow{3}{*}{$\M_G$} & V\&K & 26.0 & 26.3 & 8.2 & - & 38.0 & 31.2 & 16.7 & 12.9 & 26.4 & 30.2 & 36.6 & 25.1 & 25.2 \\
		& & R4CNN & 28.9 & 24.7 & 7.4 & - & 38.6 & 29.5 & 13.3 & 12.7 & 28.2 & 24.2 & 33.0 & 27.5 & 24.4 \\
		& & Mask-RCNN & 31.3	& 28.2	& 14.8	& -	& 42.1	& 26.0	& 26.3	& 19.9	& 35.4	& 30.5	& 44.0	& 31.4	& 30.0 \\
		\cline{2-16}
		& \multirow{3}{*}{$\M_G+$} & V\&K & 32.9 & 26.5 & 10.4 & - & 42.7 & 37.6 & 18.5 &	13.6 & 29.5 & 27.5 & 37.3 & 23.9 & 27.3 \\
		& & R4CNN & 35.0 & 25.5 & 7.7 &	- & 37.7 & 36.2 & 14.2 &	16.4 & 30.7 & 25.4 & 33.3 & 27.5 & 26.3 \\
		& & Mask-RCNN & 40.4	& 33.7	& 16.0	& -	& 49.3	& 32.1	& \textbf{29.3}	& 20.5	& \textbf{38.2}	& 31.6	& 44.3	& \textbf{33.0}	& 33.5 \\
		\hline
	\end{tabular}
	\caption{Performance of our Geodesic Bin \& Delta models under the AVP metric using detected bounding boxes. We compare with V\&K \citep{Tulsiani:CVPR15}, R4CNN \citep{Su:ICCV15} and Massa \citep{Massa:BMVC16}. We use the bounding boxes provided by V\&K \citep{Tulsiani:CVPR15} and R4CNN  \citep{Su:ICCV15}. Higher is better. Note that our models do not have any supervision on the azimuth angles unlike the other methods which use cross-entropy loss on discretized azimuth angle bins.}
	\label{table:avp_detailed}
\end{table*}

\bibliographystyle{spbasic}
\bibliography{vidal,recognition,geometry}

\begin{thebibliography}{47}
\providecommand{\natexlab}[1]{#1}
\providecommand{\url}[1]{{#1}}
\providecommand{\urlprefix}{URL }
\expandafter\ifx\csname urlstyle\endcsname\relax
  \providecommand{\doi}[1]{DOI~\discretionary{}{}{}#1}\else
  \providecommand{\doi}{DOI~\discretionary{}{}{}\begingroup
  \urlstyle{rm}\Url}\fi
\providecommand{\eprint}[2][]{\url{#2}}

\bibitem[{Aubry et~al(2014)Aubry, Maturana, Efros, Russell, and
  Sivic}]{Aubry:CVPR14}
Aubry M, Maturana D, Efros AA, Russell BC, Sivic J (2014) Seeing {3D} {C}hairs:
  {E}xemplar {P}art-{B}ased {2D}-{3D} {A}lignment {U}sing a {L}arge {D}ataset
  of {CAD} {M}odels. In: 2014 IEEE Conference on Computer Vision and Pattern
  Recognition, pp 3762--3769, \doi{10.1109/CVPR.2014.487}

\bibitem[{Chatfield et~al(2014)Chatfield, Simonyan, Vedaldi, and
  Zisserman}]{Chatfield:BMVC14}
Chatfield K, Simonyan K, Vedaldi A, Zisserman A (2014) Return of the {D}evil in
  the {D}etails: {D}elving {D}eep into {C}onvolutional {N}ets. In: British
  Machine Vision Conference

\bibitem[{Crivellaro et~al(2015)Crivellaro, Rad, Verdie, Yi, Fua, and
  Lepetit}]{Crivellaro:ICCV15}
Crivellaro A, Rad M, Verdie Y, Yi KM, Fua P, Lepetit V (2015) A {N}ovel
  {R}epresentation of {P}arts for {A}ccurate {3D} {O}bject {D}etection and
  {T}racking in {M}onocular {I}mages. In: 2015 IEEE International Conference on
  Computer Vision (ICCV), pp 4391--4399, \doi{10.1109/ICCV.2015.499}

\bibitem[{Deng et~al(2009)Deng, Dong, Socher, Li, Li, and Fei-Fei}]{ImageNet}
Deng J, Dong W, Socher R, Li LJ, Li K, Fei-Fei L (2009) Imagenet: A large-scale
  hierarchical image database. In: 2009 IEEE Conference on Computer Vision and
  Pattern Recognition, pp 248--255, \doi{10.1109/CVPR.2009.5206848}

\bibitem[{Elhoseiny et~al(2016)Elhoseiny, El-Gaaly, Bakry, and
  Elgammal}]{Elhoseiny:ICML16}
Elhoseiny M, El-Gaaly T, Bakry A, Elgammal A (2016) A {C}omparative {A}nalysis
  and {S}tudy of {M}ultiview {CNN} {M}odels for {J}oint {O}bject
  {C}ategorization and {P}ose {E}stimation. In: Proceedings of the 33rd
  International Conference on International Conference on Machine Learning -
  Volume 48, JMLR.org, ICML'16, pp 888--897

\bibitem[{Everingham et~al(2015)Everingham, Eslami, Van~Gool, Williams, Winn,
  and Zisserman}]{PASCAL}
Everingham M, Eslami SMA, Van~Gool L, Williams CKI, Winn J, Zisserman A (2015)
  The {P}ascal {V}isual {O}bject {C}lasses {C}hallenge: {A} {R}etrospective.
  International Journal of Computer Vision 111(1):98--136

\bibitem[{Girshick(2015)}]{Girshick:ICCV15}
Girshick R (2015) Fast {R-CNN}. In: 2015 IEEE International Conference on
  Computer Vision (ICCV), pp 1440--1448, \doi{10.1109/ICCV.2015.169}

\bibitem[{Girshick et~al(2018)Girshick, Radosavovic, Gkioxari, Doll\'{a}r, and
  He}]{Detectron2018}
Girshick R, Radosavovic I, Gkioxari G, Doll\'{a}r P, He K (2018) Detectron.
  \url{https://github.com/facebookresearch/detectron}

\bibitem[{Glasner et~al(2011)Glasner, Galun, Alpert, Basri, and
  Shakhnarovich}]{Glasner:ICCV11}
Glasner D, Galun M, Alpert S, Basri R, Shakhnarovich G (2011) Viewpoint-aware
  object detection and pose estimation. In: 2011 International Conference on
  Computer Vision, pp 1275--1282, \doi{10.1109/ICCV.2011.6126379}

\bibitem[{Grabner et~al(2018)Grabner, Roth, and Lepetit}]{Grabner:CVPR18}
Grabner A, Roth PM, Lepetit V (2018) {3D Pose Estimation and 3D Model Retrieval
  for Objects in the Wild}. In: {IEEE} Conference on Computer Vision and
  Pattern Recognition

\bibitem[{G{\"{u}}ler et~al(2017)G{\"{u}}ler, Trigeorgis, Antonakos, Snape,
  Zafeiriou, and Kokkinos}]{Guler:CVPR17}
G{\"{u}}ler RA, Trigeorgis G, Antonakos E, Snape P, Zafeiriou S, Kokkinos I
  (2017) {DenseReg: Fully Convolutional Dense Shape Regression In-the-Wild}.
  In: 2017 IEEE Conference on Computer Vision and Pattern Recognition (CVPR),
  pp 2614--2623, \doi{10.1109/CVPR.2017.280}

\bibitem[{G{\"{u}}ler et~al(2018)G{\"{u}}ler, Trigeorgis, Antonakos, Snape,
  Zafeiriou, and Kokkinos}]{Guler:arxiv18}
G{\"{u}}ler RA, Trigeorgis G, Antonakos E, Snape P, Zafeiriou S, Kokkinos I
  (2018) {DenseReg: Fully Convolutional Dense Shape Regression In-the-Wild}.
  coRR abs/180302188

\bibitem[{Hartley and Zisserman(2004)}]{Hartley-Zisserman04}
Hartley R, Zisserman A (2004) Multiple View Geometry in Computer Vision, 2nd
  edn. Cambridge

\bibitem[{He et~al(2016{\natexlab{a}})He, Zhang, Ren, and Sun}]{He:CVPR16}
He K, Zhang X, Ren S, Sun J (2016{\natexlab{a}}) {Deep Residual Learning for
  Image Recognition}. In: 2016 IEEE Conference on Computer Vision and Pattern
  Recognition (CVPR), pp 770--778, \doi{10.1109/CVPR.2016.90}

\bibitem[{He et~al(2016{\natexlab{b}})He, Zhang, Ren, and Sun}]{He:ECCV16}
He K, Zhang X, Ren S, Sun J (2016{\natexlab{b}}) Identity mappings in deep
  residual networks. In: Leibe B, Matas J, Sebe N, Welling M (eds) Computer
  Vision -- ECCV 2016, Springer International Publishing, Cham, pp 630--645

\bibitem[{He et~al(2017)He, Gkioxari, Dollar, and Girshick}]{He:ICCV17}
He K, Gkioxari G, Dollar P, Girshick R (2017) {Mask R-CNN}. In: 2017 IEEE
  International Conference on Computer Vision (ICCV), pp 2980--2988,
  \doi{10.1109/ICCV.2017.322}

\bibitem[{Hejrati and Ramanan(2012)}]{Hejrati:NIPS12}
Hejrati M, Ramanan D (2012) {Analyzing 3D Objects in Cluttered Images}. In:
  Pereira F, Burges CJC, Bottou L, Weinberger KQ (eds) Advances in Neural
  Information Processing Systems 25, Curran Associates, Inc., pp 593--601

\bibitem[{Hejrati and Ramanan(2014)}]{Hejrati:CVPR14}
Hejrati M, Ramanan D (2014) {Analysis by Synthesis: 3D Object Recognition by
  Object Reconstruction}. In: 2014 IEEE Conference on Computer Vision and
  Pattern Recognition, pp 2449--2456, \doi{10.1109/CVPR.2014.314}

\bibitem[{Hou et~al(2018)Hou, Miolane, Khanal, Lee, Alansary, McDonagh, Hajnal,
  Rueckert, Glocker, and Kainz}]{Hou:arxiv18}
Hou B, Miolane N, Khanal B, Lee MC, Alansary A, McDonagh S, Hajnal JV, Rueckert
  D, Glocker B, Kainz B (2018) {Computing CNN Loss and Gradients for Pose
  Estimation with Riemannian Geometry}. coRR abs/180501026

\bibitem[{Jordan and Jacobs(1994)}]{Jordan:NC94}
Jordan MI, Jacobs RA (1994) {Hierarchical Mixtures of Experts and the EM
  Algorithm}. Neural Computation 6(2):181--214, \doi{10.1162/neco.1994.6.2.181}

\bibitem[{Kendall and Cipolla(2016)}]{Kendall:ICRA16}
Kendall A, Cipolla R (2016) {Modelling uncertainty in deep learning for camera
  relocalization}. In: 2016 IEEE International Conference on Robotics and
  Automation (ICRA), pp 4762--4769, \doi{10.1109/ICRA.2016.7487679}

\bibitem[{Kendall and Cipolla(2017)}]{Kendall:CVPR17}
Kendall A, Cipolla R (2017) {Geometric Loss Functions for Camera Pose
  Regression with Deep Learning}. In: 2017 IEEE Conference on Computer Vision
  and Pattern Recognition (CVPR), pp 6555--6564, \doi{10.1109/CVPR.2017.694}

\bibitem[{Kendall et~al(2015)Kendall, Grimes, and Cipolla}]{Kendall:ICCV15}
Kendall A, Grimes M, Cipolla R (2015) Posenet: A convolutional network for
  real-time 6-dof camera relocalization. In: 2015 IEEE International Conference
  on Computer Vision (ICCV), pp 2938--2946, \doi{10.1109/ICCV.2015.336}

\bibitem[{Li et~al(2018)Li, Bai, and Hager}]{Li:arxiv18}
Li C, Bai J, Hager GD (2018) {A Unified Framework for Multi-View Multi-Class
  Object Pose Estimation}. coRR abs/180108103

\bibitem[{Liebelt and Schmid(2010)}]{Liebelt:CVPR10}
Liebelt J, Schmid C (2010) {Multi-view object class detection with a 3D
  geometric model}. In: 2010 IEEE Computer Society Conference on Computer
  Vision and Pattern Recognition, pp 1688--1695,
  \doi{10.1109/CVPR.2010.5539836}

\bibitem[{Lim et~al(2013)Lim, Pirsiavash, and Torralba}]{Lim:ICCV13}
Lim JJ, Pirsiavash H, Torralba A (2013) Parsing ikea objects: Fine pose
  estimation. In: 2013 IEEE International Conference on Computer Vision, pp
  2992--2999, \doi{10.1109/ICCV.2013.372}

\bibitem[{Lim et~al(2014)Lim, Khosla, and Torralba}]{Lim:ECCV14}
Lim JJ, Khosla A, Torralba A (2014) {FPM: Fine Pose Parts-Based Model with 3D
  CAD Models}. In: Fleet D, Pajdla T, Schiele B, Tuytelaars T (eds) Computer
  Vision -- ECCV 2014, Springer International Publishing, Cham, pp 478--493

\bibitem[{López-Sastre et~al(2011)López-Sastre, Tuytelaars, and
  Savarese}]{Sastre:ICCVW11}
López-Sastre RJ, Tuytelaars T, Savarese S (2011) {Deformable part models
  revisited: A performance evaluation for object category pose estimation}. In:
  2011 IEEE International Conference on Computer Vision Workshops (ICCV
  Workshops), pp 1052--1059, \doi{10.1109/ICCVW.2011.6130367}

\bibitem[{Ma et~al(2003)Ma, Soatto, Kosecka, and Sastry}]{MASKS03}
Ma Y, Soatto S, Kosecka J, Sastry S (2003) An Invitation to 3D Vision: From
  Images to Geometric Models. Springer Verlag

\bibitem[{Mahendran et~al(2017)Mahendran, Ali, and Vidal}]{Mahendran:ICCVW17}
Mahendran S, Ali H, Vidal R (2017) {3D Pose Regression Using Convolutional
  Neural Networks}. In: 2017 IEEE International Conference on Computer Vision
  Workshop (ICCVW), vol~00, pp 2174--2182, \doi{10.1109/ICCVW.2017.254}

\bibitem[{Mahendran et~al(2018)Mahendran, Ali, and Vidal}]{Mahendran:arxiv18}
Mahendran S, Ali H, Vidal R (2018) {A Mixed Classification-Regression Framework
  for 3D Pose Estimation from 2D Images}. coRR abs/180503225

\bibitem[{Massa et~al(2014)Massa, Aubry, and Marlet}]{Massa:arxiv14}
Massa F, Aubry M, Marlet R (2014) {Convolutional Neural Networks for joint
  object detection and pose estimation: A comparative study}. CoRR abs/14127190

\bibitem[{Massa et~al(2016)Massa, Marlet, and Aubry}]{Massa:BMVC16}
Massa F, Marlet R, Aubry M (2016) Crafting a multi-task {CNN} for viewpoint
  estimation. In: British Machine Vision Conference

\bibitem[{Mousavian et~al(2017)Mousavian, Anguelov, Flynn, and
  Košecká}]{Mousavian:CVPR17}
Mousavian A, Anguelov D, Flynn J, Košecká J (2017) {3D Bounding Box
  Estimation Using Deep Learning and Geometry}. In: 2017 IEEE Conference on
  Computer Vision and Pattern Recognition (CVPR), pp 5632--5640,
  \doi{10.1109/CVPR.2017.597}

\bibitem[{Pavlakos et~al(2017)Pavlakos, Zhou, Chan, Derpanis, and
  Daniilidis}]{Pavlakos:ICRA17}
Pavlakos G, Zhou X, Chan A, Derpanis KG, Daniilidis K (2017) 6-dof object pose
  from semantic keypoints. In: 2017 IEEE International Conference on Robotics
  and Automation (ICRA), pp 2011--2018, \doi{10.1109/ICRA.2017.7989233}

\bibitem[{Pepik et~al(2012{\natexlab{a}})Pepik, Gehler, Stark, and
  Schiele}]{Pepik:ECCV12}
Pepik B, Gehler P, Stark M, Schiele B (2012{\natexlab{a}}) {3D2PM -- 3D
  Deformable Part Models}. In: Fitzgibbon A, Lazebnik S, Perona P, Sato Y,
  Schmid C (eds) Computer Vision -- ECCV 2012, Springer Berlin Heidelberg,
  Berlin, Heidelberg, pp 356--370

\bibitem[{Pepik et~al(2012{\natexlab{b}})Pepik, Stark, Gehler, and
  Schiele}]{Pepik:CVPR12}
Pepik B, Stark M, Gehler P, Schiele B (2012{\natexlab{b}}) {Teaching 3D
  geometry to deformable part models}. In: 2012 IEEE Conference on Computer
  Vision and Pattern Recognition, pp 3362--3369,
  \doi{10.1109/CVPR.2012.6248075}

\bibitem[{Rad and Lepetit(2017)}]{Rad:ICCV17}
Rad M, Lepetit V (2017) {BB8: A Scalable, Accurate, Robust to Partial Occlusion
  Method for Predicting the 3D Poses of Challenging Objects without Using
  Depth}. In: 2017 IEEE International Conference on Computer Vision (ICCV), pp
  3848--3856, \doi{10.1109/ICCV.2017.413}

\bibitem[{Savarese and Fei-Fei(2007)}]{Savarese:ICCV07}
Savarese S, Fei-Fei L (2007) {3D generic object categorization, localization
  and pose estimation}. In: 2007 IEEE 11th International Conference on Computer
  Vision, pp 1--8, \doi{10.1109/ICCV.2007.4408987}

\bibitem[{Savarese and Fei-Fei(2008)}]{Savarese:ECCV08}
Savarese S, Fei-Fei L (2008) {View Synthesis for Recognizing Unseen Poses of
  Object Classes}. In: Forsyth D, Torr P, Zisserman A (eds) Computer Vision --
  ECCV 2008, Springer Berlin Heidelberg, Berlin, Heidelberg, pp 602--615

\bibitem[{Su et~al(2015)Su, Qi, Li, and Guibas}]{Su:ICCV15}
Su H, Qi CR, Li Y, Guibas LJ (2015) {Render for CNN: Viewpoint Estimation in
  Images Using CNNs Trained with Rendered 3D Model Views}. In: 2015 IEEE
  International Conference on Computer Vision (ICCV), pp 2686--2694,
  \doi{10.1109/ICCV.2015.308}

\bibitem[{Tulsiani and Malik(2015)}]{Tulsiani:CVPR15}
Tulsiani S, Malik J (2015) {Viewpoints and keypoints}. In: 2015 IEEE Conference
  on Computer Vision and Pattern Recognition (CVPR), pp 1510--1519,
  \doi{10.1109/CVPR.2015.7298758}

\bibitem[{Tulsiani et~al(2018)Tulsiani, Gupta, Fouhey, Efros, and
  Malik}]{Tulsiani:CVPR17}
Tulsiani S, Gupta S, Fouhey D, Efros AA, Malik J (2018) {Factoring Shape, Pose,
  and Layout from the 2D Image of a 3D Scene}. In: {IEEE} Conference on
  Computer Vision and Pattern Recognition

\bibitem[{Wang et~al(2016)Wang, Li, Jia, and Liang}]{Wang:PCM16}
Wang Y, Li S, Jia M, Liang W (2016) Viewpoint estimation for objects with
  convolutional neural network trained on synthetic images. In: Chen E, Gong Y,
  Tie Y (eds) Advances in Multimedia Information Processing - PCM 2016,
  Springer International Publishing, Cham, pp 169--179

\bibitem[{Wang et~al(2018)Wang, Tan, Yang, Liu, Ding, Zhou, and
  Davis}]{Wang:arxiv18}
Wang Y, Tan X, Yang Y, Liu X, Ding E, Zhou F, Davis LS (2018) {3D Pose
  Estimation for Fine-Grained Object Categories}. coRR abs/180604314

\bibitem[{Wu et~al(2016)Wu, Xue, Lim, Tian, Tenenbaum, Torralba, and
  Freeman}]{Wu:ECCV16}
Wu J, Xue T, Lim JJ, Tian Y, Tenenbaum JB, Torralba A, Freeman WT (2016)
  {Single Image 3D Interpreter Network}. In: Leibe B, Matas J, Sebe N, Welling
  M (eds) Computer Vision -- ECCV 2016, Springer International Publishing,
  Cham, pp 365--382

\bibitem[{Xiang et~al(2014)Xiang, Mottaghi, and Savarese}]{Xiang:WACV14}
Xiang Y, Mottaghi R, Savarese S (2014) {Beyond PASCAL: A benchmark for 3D
  object detection in the wild}. In: IEEE Winter Conference on Applications of
  Computer Vision, pp 75--82, \doi{10.1109/WACV.2014.6836101}

\end{thebibliography}
\end{document}